\documentclass[10pt,journal,compsoc]{IEEEtran}
\ifCLASSOPTIONcompsoc
  \usepackage[nocompress]{cite}
\else
  \usepackage{cite}
\fi
\ifCLASSINFOpdf
\else
\fi
\usepackage{amsmath,amsfonts}
\usepackage{algorithmic}
\usepackage{algorithm}
\usepackage{array}
\usepackage[caption=false,font=normalsize,labelfont=sf,textfont=sf]{subfig}
\usepackage{textcomp}
\usepackage{stfloats}
\usepackage{url}
\usepackage{verbatim}
\usepackage{graphicx}
\usepackage{multirow}
\usepackage{multicol}
\usepackage{makecell}
\usepackage{color}
\usepackage{pifont}

\usepackage{colortbl}
\usepackage{amsthm}
\usepackage{amssymb}
\usepackage{booktabs}

\usepackage{mathrsfs}
\usepackage{xcolor}
\usepackage{ragged2e}

\newtheorem{theorem}{Theorem}

\DeclareMathOperator{\argmin}{argmin}
\DeclareMathOperator{\argmax}{argmax}

\newcommand{\tabincell}[2]{\begin{tabular}{@{}#1@{}}#2\end{tabular}}

\usepackage{tikz}
\usetikzlibrary{shadings}
\newcommand*{\Mt}[2]{ \protect\tikz[align=center] {\protect\node[shape=circle,fill=#1, scale=0.7, align=center,line width=0.5pt, draw=black](X){#2};}}

\newcommand{\eg}{\emph{e.g.}}
\newcommand{\ie}{\emph{i.e.}}

\begin{document}
%
\title{Divert More Attention to Vision-Language \\ Object Tracking}

\author{
Mingzhe Guo$^{*}$, Zhipeng Zhang$^{*\dagger}$, Liping Jing$^{\dagger}$, Haibin Ling,~\IEEEmembership{Fellow,~IEEE,} and Heng Fan
\\
\IEEEcompsocitemizethanks{
\IEEEcompsocthanksitem $^{*}$Equal contributions. $^{\dagger}$Corresponding author. 
\IEEEcompsocthanksitem This work is co-supervised by Prof. Liping Jing and Dr. Zhipeng Zhang.
\IEEEcompsocthanksitem A preliminary version\cite{vlt} of this work has appeared in NeurIPS 2022.
}
}

\markboth{Submitted to IEEE Journal}%
{Shell \MakeLowercase{\textit{et al.}}: Bare Demo of IEEEtran.cls for Computer Society Journals}

\IEEEtitleabstractindextext{%

\begin{abstract}
\justifying Multimodal vision-language (VL) learning has noticeably pushed the tendency toward generic intelligence owing to emerging large foundation models. However, tracking, as a fundamental vision problem, surprisingly enjoys less bonus from recent flourishing VL learning. We argue that the reasons are two-fold: the lack of large-scale vision-language annotated videos and ineffective vision-language interaction learning of current works. These nuisances motivate us to design more effective vision-language representation for tracking, meanwhile constructing a large database with language annotation for model learning. Particularly, in this paper, we first propose a general attribute annotation strategy to decorate videos in six popular tracking benchmarks, which contributes a large-scale vision-language tracking database with more than 23,000 videos. We then introduce a novel framework to improve tracking by learning a unified-adaptive VL representation, where the cores are the proposed asymmetric architecture search and modality mixer (ModaMixer). To further improve VL representation, we introduce a contrastive loss to align different modalities. To thoroughly evidence the effectiveness of our method, we integrate the proposed framework on three tracking methods with different designs, \emph{i.e.,} the CNN-based SiamCAR~\cite{SiamCAR}, the Transformer-based OSTrack~\cite{ostrack}, and the hybrid structure TransT~\cite{TransT}. The experiments demonstrate that our framework can significantly improve all baselines on six benchmarks. Besides empirical results, we theoretically analyze our approach to show its rationality. By revealing the potential of VL representation, we expect the community to divert more attention to VL tracking and hope to open more possibilities for future tracking with diversified multimodal messages. 

\end{abstract}

}

\maketitle

\IEEEdisplaynontitleabstractindextext

\IEEEpeerreviewmaketitle

\IEEEraisesectionheading{\section{Introduction}\label{intro}}

\IEEEPARstart{M}{ultimodal} vision-language (VL) learning is initially treated as an independent research field, a surge of downstream tasks, like visual question answering (VQA) ~\cite{vlvqa1,vlvqa2} and image/video caption~\cite{vlcaption1,vlcaption2}, have drawn great attention for years. The fundamental vision tasks, like object detection~\cite{detsurvey} and segmentation~\cite{segsurvey}, walk on a different path, which focuses on how to construct powerful deep networks for pure visual input. The separation did not change until the release of CLIP~\cite{clip}, and recently took a stride forward with the emergence of segment anything model~\cite{sam} and VisionLLM~\cite{visionllm}. In recent works, language description usually serves as instructions for the basic model to build its output. Previous works, \emph{e.g.,}~\cite{SNLT},~\cite{Li},~\cite{Feng}, attempt to introduce language as a complement to the image template. However, owing to the lack of language-annotated databases and sophisticated model design, it still needs deeper study to unleash the power of vision-language tracking.

One crucial reason attributed to the rapid development of visual object tracking is the introduced benchmarks in recent years, \emph{e.g.,} OTB~\cite{OTB2015}, TrackingNet~\cite{trackingnet}, LaSOT~\cite{LaSOT}, GOT-10k~\cite{GOT10K}, TNL2K~\cite{TNL2K}, however, among which only 2,471 videos are annotated with language description. Besides the small amount, it is noticed that the language description of a target is formulated as a sentence, \emph{e.g.,} ``black bicycle by a man on the mountain road''. This raises the risk that the model may overfit a specific language style, eventually degrading the performance and generalization of the tracking model. Therefore, a larger database with a general language annotation style is urgent to facilitate the study of vision-language tracking.

\begin{figure}[!t]
\centering
\includegraphics[width=\linewidth]{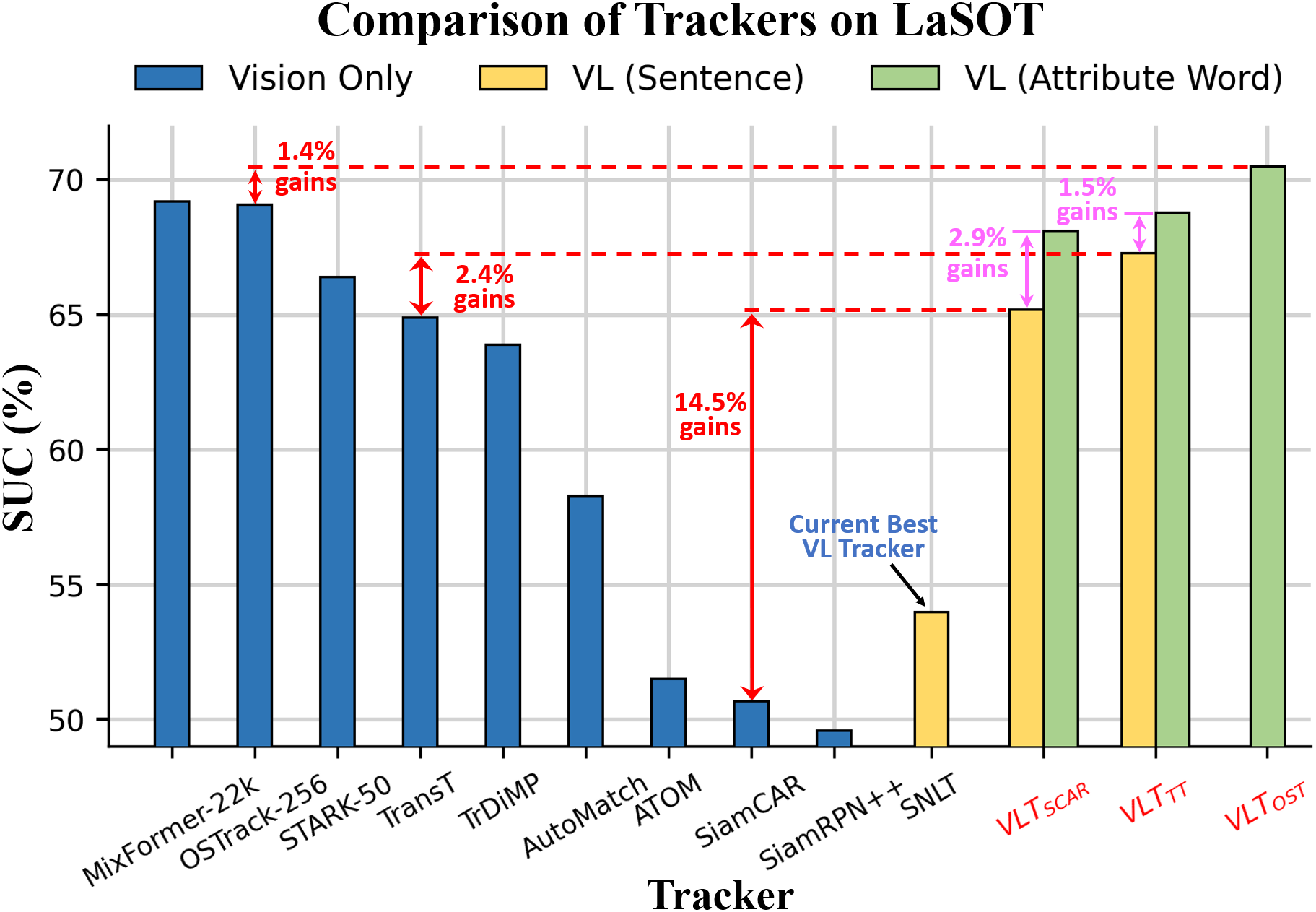}  
\vspace{-10pt}
\caption{
Comparison between vision-only and multimodal vision-language (VL) trackers on LaSOT~\cite{LaSOT}.
}
\label{fig:sota_comp}
\end{figure}

In addition to training data, the quality of the learned VL representation is another essential factor for constructing a robust tracking model. The previous works~\cite{SNLT,Li,Feng} consider language description of the target as a prompt (template) to search the pre-defined object in the visual features of the search region. Despite the simplicity, their performances fall much short of current state-of-the-art methods (see Fig.~\ref{fig:sota_comp}). We blame the backwardness on the following
reasons: (a) vision and language are {independently} treated and processed {distantly} until the final matching module. Although this manner can easily connect the two modalities, it does not accord with human learning procedure that integrates multisensory by various neurons before causal inference~\cite{pnas_causal}, which inevitably results in a lower upper bound for VL tracking. {{A more unified method is desired for deep interaction of vision and language}}. (b)  template and search branches are treated as {homoplasmic} inputs, and previous methods adopt symmetric feature learning structures for these two branches, inherited from typical vision-only Siamese tracking~\cite{SNLT}. We argue the mixed modality may have different intrinsic nature than the pure vision modality, and thus {{requires a more flexible and adaptive design for different signals}}.

Having observed above, we first establish a large-scale VL database, which describes the target with separate words instead of a sentence as in the previous benchmarks~\cite{LaSOT, TNL2K, Li}. Specifically, each language annotation contains four attribute words, including major class (\textit{e.g.,} bicycle), root class (\textit{e.g.,} vehicle), color (\textit{e.g.,} green), and the initial position of the target (\textit{e.g.,} center). Intuitively, each attribute word provides high-level linguistic information about the target from different perspectives, which is believed to be helpful for distinguishing the target from the background and other objects. The intrinsic motivation behind our design is that attribute words naturally decrease the risk brought by different sentence annotation styles, meanwhile avoiding the negative influence of redundant prepositions and articles. We annotate six popular tracking datasets (i.e., LaSOT~\cite{LaSOT}, LaSOT-Extension~\cite{LaSOT_Extention}, TNL2K~\cite{TNL2K}, GOT-10K~\cite{GOT10K}, TrackingNet~\cite{trackingnet} and OTB99-Language~\cite{Li}) for more than 23,000 videos, which is orders of scale larger than previous VL-labeled benchmarks. Moreover, we experimentally prove that the proposed attribute-based annotation strategy is more effective than the de facto sentence-based labeling style for training vision-language tracking models.

Then, we propose Modality Mixer (ModaMixer) and Asymmetric Searching Strategy (ASearch) to learn vision-language representation. Particularly, ModaMixer is a conceptually simple yet effective module for VL interaction, which is inspired by the common sense that visual channel features reveal object semantics as in language~\cite{channelmeaning1,channelmeaning2}. ModaMixer regards language representation as a selector to reweight different channels of visual features, enhancing target-specific channels as well as suppressing irrelevant ones. The selected feature is then fused with the original visual feature, using a special asymmetric design (analyzed later), to generate the final unified VL representation. Furthermore, we apply the contrastive loss~\cite{clip} to ModaMixer during training to promote the alignment between language and visual representations. ModaMixers are integrated into different stages of the backbone network to boost the robustness and discriminability of the unified VL representation at different semantic levels. 

Besides the fusion strategy, the tracking model's basic structure is another important factor affecting VL representation quality. Classical methods adopt the Siamese framework since their inputs are pure visual images. When injecting language information into vision features, the priori hypotheses of symmetric model design may not be optimal. In this work, we propose an asymmetric searching strategy (ASearch) to adaptively learn the model structure. ASearch borrows the idea from neural architecture search (NAS)~\cite{nas1,nas2} to separately learn {distinctive} and {asymmetric} networks for mixed modality in different branches and ModaMixers. Notably, our proposed ASearch avoids burdensome re-training on ImageNet~\cite{imagenet}, enabling quick reproducibility of our work (only 0.625 GPU days with a single RTX-2080Ti). Our ASearch is general and flexible, and together with ModaMixer, it could benefit trackers of different architectures with the learned unified-adaptive VL representation.

To demonstrate the effectiveness and generality of our framework, we apply it to different tracking methods (\textit{i.e.,} SiamCAR~\cite{SiamCAR}, TransT~\cite{TransT} and OSTrack~\cite{ostrack}) and train them with the introduced database. As shown in Fig.~\ref{fig:sota_comp}, the proposed VLT series outperforms the baselines by a considerable margin, which surprised us most is that the proposed framework improves the performance of pure-CNN baseline SiamCAR~\cite{SiamCAR} to the level of Transformer tracker.

In summary, we make the following contributions: 
\begin{itemize}
    \item We propose the attribute annotation strategy and annotate six popular tracking datasets with general attribute words, providing a strong database for VL tracking.

    \item We propose a conceptually simple yet novel and effective Modality Mixer (ModaMixer) for unified VL representation learning.

    \item We present the asymmetric searching strategy (ASearch) that adapts the mixed VL representation for improved tracking robustness.

    \item With multiple baselines of different architectures, our VL tracking framework generally improves the performance with considerable gains.
\end{itemize}

Notably, this paper is an extension of our previous work~\cite{vlt}, including new contributions as follows: \textbf{(i)} we provide more than 23,000 VL videos annotated with general attribute words for VL representation learning; \textbf{(ii)} we expand the ModaMixer with contrastive loss training; \textbf{(iii)} we apply our method on one-stream OSTrack and reach a new SOTA level, and \textbf{(iv)} we provide more analyses on our method and different language annotations. 

Our source code, models and database will be released at~\url{https://github.com/JudasDie/SOTS}.

\begin{figure*}[t]
\centering
\includegraphics[width =\textwidth]{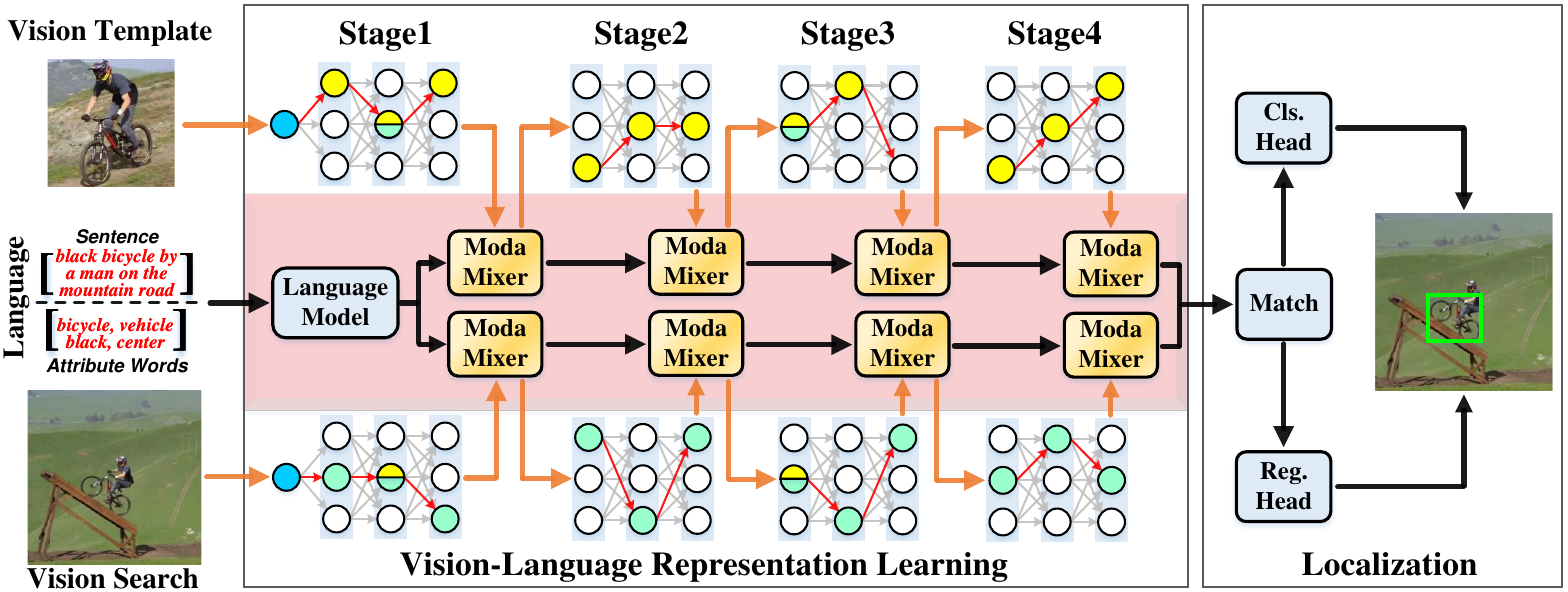}
\caption{The proposed vision-language tracking framework. The semantic information of language description (\ie, sentence or attribute words) is injected to vision from shallow to deep layers of the asymmetric modeling architecture to learn unified-adaptive vision-language representation.}
\label{fig:framework}
\end{figure*}

\section{Related Work}
{\noindent \textbf{Visual Tracking.}} Tracking has witnessed great progress in the past decades. One representative branch is tracking with correlation filter (CF), which solves the object template matching problem by the convolution theorem. The seminal work MOSSE~\cite{MOSSE} first introduces the correlation operation in the frequency domain into visual tracking, motivating a series of CF-based works~\cite{KCF,ECO,Atom,prdimp}. Siamese tracking~\cite{Siamfc,tao2016siamese} is another popular branch for well-balanced tracking accuracy and efficiency, and has revolutionized with numerous extensions on a pure-CNN architecture~\cite{Siamrpn,SiamDW,Siamrpn++,C-RPN,Ocean}. Recently, Transformer~\cite{Transformer} has been introduced to Siamese tracking for better interactions of visual features and greatly pushed the standard of state-of-the-art performance~\cite{TransT,TrDiMP,stark}. In particular, the uprising pure-Transformer one-stream trackers~\cite{ostrack,mixformer} achieve the best results by jointly modeling the template and search tokens. In this paper, we choose Siamese tracking as the base architecture for its effectiveness and efficiency. Different from the previous works inheriting the symmetric modeling structure, we search an asymmetric network to better unleash the intrinsic nature of each branch.

{\noindent \textbf{Vision-Language Tracking.}} Natural language contains high-level semantics and has been leveraged to foster vision-related tasks~\cite{languagetask1,languagetask2,languagetask3} including tracking~\cite{Li,Feng,SNLT}. The work~\cite{Li} first introduces linguistic description to tracking and shows that language can enhance the robustness of vision-based method. Most recently, SNLT~\cite{SNLT} considers language description as an additional prompt besides the visual template, and shows that the complementary clues between the two modalities can benefit feature matching ability of Siamese tracking methods.  Despite its improvement, VL tracking is still limited by insufficient datasets with circumscribed sentence annotation and weak connections between VL modalities. To unleash the power of VL tracking, we present a general database for effective feature learning and a novel framework to learn a unified-adaptive vision-language representation.

{\noindent \textbf{Vision-Language in Other Tasks.}} With a wave of progress in multimodal foundation models (\eg, CLIP~\cite{clip} and CoCa~\cite{coca}), many natural vision-language tasks benefit from the language supervision, including image captioning~\cite{imgcaption1,imgcaption2}, visual grounding~\cite{visground1,visground2}, \textit{etc.} Classical visual perception tasks, \eg, image classification~\cite{vlclassification}, object detection~\cite{vldetection}, referring segmentation~\cite{vlsegmentation}, also enjoy the bonus by introducing external language modality. Recently, the emergence of large visual foundation models (\eg, SAM~\cite{sam} and VisionLLM~\cite{visionllm}) raises more attentions on multimodal VL learning, which generally regards language as a prompt to instruct downstream tasks. From a different perspective, we  explore the vision-language relation and unify both to learn a stronger VL representation for tracking.

{\noindent \textbf{NAS for Tracking.}} Neural architecture search (NAS) aims at finding the optimal design of deep network architectures~\cite{nas1,nas2,DARTS,spos} and has been introduced to tracking~\cite{yan2021lighttrack,AutoMatch}. LightTrack~\cite{yan2021lighttrack} tends to search a lightweight backbone but is computationally demanding (about 40 V100 GPU days). AutoMatch uses DARTS~\cite{DARTS} to find better matching networks for Siamese tracking. All these methods leverage NAS for vision-only tracking and search a {symmetric} Siamese architecture. Differently, our work searches the network for multimodal tracking and tries to find a more general and flexible {asymmetric} two-stream counterpart. In addition, our search pipeline only takes 0.625 RTX-2080Ti GPU days, which is much more resource-friendly.

\section{Approach and Database for Unified-Adaptive Vision-Language Tracking}

This section details our proposed approach and database for unified-adaptive vision-language (VL) tracking, as shown in Fig.~\ref{fig:framework}. In specific, we first describe the attribute annotation strategy to provide a strong database for VL tracking. Then, we present the proposed modality mixer for generating unified multimodal representation and the asymmetric searching strategy which searches the optimal model structure for learning adaptive VL representation. Afterwards, we illustrate the proposed tracking framework, followed by theoretical analysis of our method.

\begin{table}[!t]
  \caption{Summary of current popular tracking benchmarks. $\surd$ indicates the dataset is annotated with sentence/attribute description. Notably, TackingNet~\cite{trackingnet} is a high-quality subset of {YouTube-BB}~\cite{youtube}, so we only annotate the former one. Besides, to relieve the negative influence of inaccurate weighted-average box annotation in TrackingNet, we only consider the first four training splits (each split contains 2,511 videos, please refer to~\cite{trackingnet} for more details).}
  \label{tab:vldataset}
  \begin{center}
  \newcommand{\dist}{\hspace{3pt}}
  \newcommand{\best}[1]{{\textcolor{red}{#1}}}
  \newcommand{\scnd}[1]{{\textcolor{blue}{#1}}}
  \renewcommand\arraystretch{1.2}
  
  \resizebox{\linewidth}{!}{
  \begin{tabular}{c|c|c|c|c}
  \hline

  \hline
  \multirow{2}{*}{{Dataset}} &\multicolumn{2}{c|}{{Number of Videos}}  &\multicolumn{2}{c}{{Language Description}}  \\
  \cline{2-5}
  &Train &Evaluation &{Sentence} &{Attr. Word} \\

  \hline
  
  \hline
{YouTube-BB}~\cite{youtube}  &240,000 &- &- &-         \\
{TrackingNet}~\cite{trackingnet} &10,044 &511 &- &\textbf{$\surd$}         \\
  {GOT-10K}~\cite{GOT10K}  &9,335 &360 &- &\textbf{$\surd$}         \\
{LaSOT}~\cite{LaSOT}  &1,120 &280 &$\surd$ &\textbf{$\surd$}         \\
{LaSOT-EXT}~\cite{LaSOT_Extention}  &- &150 &$\surd$ &\textbf{$\surd$}         \\

  {TNL2K}~\cite{GOT10K}  &1,300 &700 &$\surd$ &\textbf{$\surd$}         \\
  {OTB99-L}~\cite{Li}  &51 &48 &$\surd$ &\textbf{$\surd$}         \\

  \hline

  \hline
  \end{tabular}}
  \end{center}
  \end{table}

\subsection{Attribute Annotation Strategy}
\label{sec:attribute}

The volume and language annotation of vision-language datasets decide the quality and generality of learned VL representation. As shown in Table~\ref{tab:vldataset}, there are only 2,471 training videos with official sentence descriptions, \ie, 1120 (LaSOT~\cite{LaSOT}) + 1300 (TNL2K~\cite{TNL2K}) + 51 (OTB99-Language~\cite{Li}), resulting in insufficient VL representation learning. Besides, the different annotation styles of current sentence descriptions make the tracking model hardly learn a general VL representation, causing severe performance degradation during inference on new videos with unofficial annotations or a different language description style. To enhance the generality, we propose the attribute annotation strategy and annotate six popular tracking datasets, providing a strong database for VL tracking.

The attribute annotation strategy describes each target of a video with
four attribute words (\ie, major class, root class, color and initial position of the target). Inspired by the success of image classification~\cite{vlclassification} and visual grounding~\cite{visground1}, we first annotate the target with simple and comprehensible class words (\ie, major class and root class), which enrich the visual clues with high-level classificatory messages. We further describe the color and initial position of the target to enhance the discernibility for the distractors of the same class (\eg, cars with different colors). Besides helpful high-level linguistic information, each attribute word is general and could be shared with other targets of the same attribute. This avoids the exclusiveness of current sentence annotation and benefits general representation learning. Finally, we annotate six popular tracking datasets (more than 23,000 videos) with proposed attribute annotation strategy (see Table~\ref{tab:vldataset}), providing a large-scale database for general VL representation learning.

\subsection{Modality Mixer for Unified Representation}
\label{sec:modalixer}

The essence of multimodal learning is a simple and effective modality fusion module. As discussed before, existing VL trackers simply treat language as a prompt to instruct visual feature learning, in which different modalities are treated independently and processed distantly until the final matching module~\cite{SNLT,Li}. Despite the effectiveness to some extent, the complementarity of different modalities in representation learning is largely underexplored, which may impede the multimodal learning to unleash its power for VL tracking. In this work, we propose the modality mixer (dubbed {ModaMixer}) to demonstrate a compact way to learn a unified VL representation for tracking.

\begin{figure}[!t]
\centering
\includegraphics[width=\linewidth]{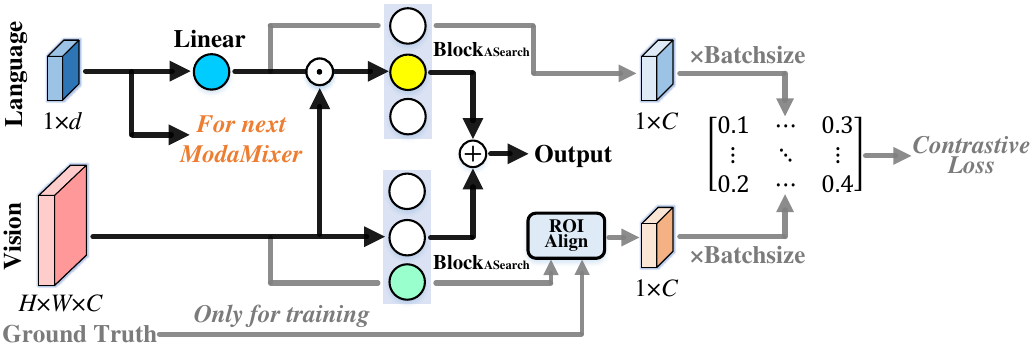}
\vspace{-11pt}
\caption{
Illustration of the ModaMixer. The inference is denoted with black arrows, and external contrastive loss~\cite{clip} during training is marked with \textcolor{black}{gray} arrows.
}
\label{fig:modamixer}
\end{figure}

ModaMixer considers language representation as a selector to reweight channels of vision features. The first step is to abstract the language description of each video\footnote{The language description in tracking is generated {only} by the initial target object in the first frame.} (\ie, one sentence or four attribute words, as shown in Fig.~\ref{fig:framework}) to semantic features with a language model~\cite{bert}. In specific, the sentence description with $N$ words of a video will be modeled as semantic features with size of $(N+2)\times d$. The extra ``2'' denotes the ``[CLS][SEP]'' characters in language processing (see~\cite{bert} for more details). Notably, the sentence descriptions for different videos may contain various length $N$. To ensure the ModaMixer applicable for all videos, we average the features for all sentence words along sequence length dimension ``(N+2)'' to generate a unique language representation $\mathbf{f}_{l}^{sent} \in \mathbb{R}^{1 \times d}$ for each sentence description. When using attribute words as the language description, we first construct the shared ``attribute representation dictionary'' by collecting each word representation $\mathbf{f}_{l}^{word} \in \mathbb{R}^{1 \times d}$, which averages the dimension ``1+2'' of corresponding semantic features with size of $(1+2)\times d$. Then the four attribute words of each target could refer to the ``dictionary'' and generate the language representation $\mathbf{f}_{l}^{attr} \in \mathbb{R}^{1 \times 4d}$ by concatenating on the channel dimension $d$.

After building the language representation, a linear layer is followed to align the channel number of $\mathbf{f}_{l}\in\{\mathbf{f}_{l}^{sent},\mathbf{f}_{l}^{attr}\}$ with the corresponding vision feature $\mathbf{f}_{v} \in \mathbb{R}^{H \times W \times C}$. Channel selector is expressed as Hadamard product operator, which point-wisely multiplies the aligned language representation ($\mathbb{R}^{1 \times C}$) to the embedding of each spatial position in the vision feature $\mathbf{f}_{v}$. Finally, a residual connection between the mixed feature $\mathbf{f}_{m}$ and vision feature $\mathbf{f}_{v}$ is conducted to avoid losing informative vision details. In a nutshell, the ModaMixer is formulated as
\begin{align}
   \mathbf{f}_{m}= \operatorname{Block_{ASearch}}{(\operatorname{Linear}(\mathbf{f}_{l}) \odot \mathbf{f}_{v})} + \operatorname{Block_{ASearch}}{(\mathbf{f}_{v})},
   \label{eq:modamixer}
\end{align}
where $\odot$ denotes Hadamard product, $\operatorname{Linear}$ is a linear projection layer with weight matrix size of $d\times C/4d\times C$ (for the language representation of sentence or attribute words, respectively) to align the channel number, and $\operatorname{Block_{ASearch}}$ indicates post-processing block before residual connection. Please note that, to enable adaptive feature modeling for different modalities, we search different $\operatorname{Block_{ASearch}}$ to process features before and after fusion (see Sec.~\ref{asymmetric} for more details). The proposed ModaMixer is illustrated in Fig.~\ref{fig:modamixer} (black arrows). Akin to the channel attention~\cite{channelattention,channelmeaning1}, the high-level semantics in language representation can dynamically enhance the target-specific channels of vision features, and meanwhile suppress the responses of irrelevant distractors belonging to both inter- and intra-classes.

Besides the fusion mechanism, we further apply the contrastive loss~\cite{clip} into ModaMixer during training (\textcolor{black}{gray} arrows in Fig.~\ref{fig:modamixer}), aiming for deeper multimodal alignment. Specifically, we first use ground truth to extract the visual target embedding $\mathbf{e}_{v} \in \mathbb{R}^{1 \times C}$ from $\operatorname{Block_{ASearch}}{(\mathbf{f}_{v})}$ with ROI align~\cite{maskrcnn}. Then, $\mathbf{e}_{v}$ and the aligned language embedding $\mathbf{e}_{l}= \operatorname{Linear}(\mathbf{f}_{l}) \in \mathbb{R}^{1 \times C}$ within a training batch of size $b$ are concatenated respectively, forming two embedding groups $\{\mathbf{e}_{v}^{1},\mathbf{e}_{v}^{2},...,\mathbf{e}_{v}^{b}\}/\{\mathbf{e}_{l}^{1},\mathbf{e}_{l}^{2},...,\mathbf{e}_{l}^{b}\} \in \mathbb{R}^{b \times C}$. Afterwards, a cosine similarity operation is performed between the two embedding groups and outputs a logit matrix $\mathbf{m}_{log} \in \mathbb{R}^{b \times b}$. Finally, the contrastive loss is computed on $\mathbf{m}_{log}$ with different modality supervision, which is formulated as
%
%
\begin{align}
   {\ell}_{contra}=[\operatorname{CE}&{(\mathbf{m}_{log},\mathbf{v}_{lbl})}+\operatorname{CE}{(\mathbf{m}_{log}^{T},\mathbf{v}_{lbl})}]/2,\\
   \mathbf{m}_{log}&=\{\mathbf{e}_{v}^{1},\mathbf{e}_{v}^{2},...,\mathbf{e}_{v}^{b}\}\otimes\{\mathbf{e}_{l}^{1},\mathbf{e}_{l}^{2},...,\mathbf{e}_{l}^{b}\} \\
   \mathbf{v}_{lbl}&=\{1,2,...,b\}
   \label{eq:clip_loss}
\end{align}
where $\operatorname{CE}$ computes the cross entropy, $\otimes$ denotes the cosine similarity and $\mathbf{v}_{lbl} \in \mathbb{R}^{1 \times b}$ is the training label (please refer to~\cite{clip} for more details). With the aligned feature space, ModaMixer can better fuse multimodal messages and learn a unified VL representation.

\begin{table*}[!t]
	\centering
	\caption{The asymmetrical architecture learned by ASearch. \Mt{cyan}{} is the stem convolution layer. (\Mt{red!70!white}{}\Mt{red!30!yellow}{}\Mt{green!60!white}{}\Mt{blue!60!white}{}) represents basic ASearch unit, where the first three ones indicate {Shuffle blocks}~\cite{shufflenet} with kernel sizes of (3,5,7), respectively, and the last one denotes a {Shuffle Xception block}~\cite{shufflenet} with kernel size 3.}
	\label{tab:framework}%
	\newcommand{\best}[1]{{\textcolor{red}{#1}}}
	\newcommand{\scnd}[1]{{\textcolor{blue}{#1}}}
	\newcommand{\opt}[1]{{\textcolor{violet}{#1}}}
	\newcommand{\fast}[1]{{\textcolor{orange}{#1}}}
	\renewcommand\arraystretch{1.2}
	\resizebox{0.9\linewidth}{!}{%
        \begin{tabular}{l|c|c|c|c|c|c|c|c|c}
        	\hline
        	           & {Stem} &{Stage1} &{\tabincell{c}{Moda\\Mixer}} &{Stage2} &{\tabincell{c}{Moda\\Mixer}} &{Stage3} &{\tabincell{c}{Moda\\Mixer}} &{Stage4} &{\tabincell{c}{Moda\\Mixer}}\\
        	\hline
        	
        	{Template}   &\Mt{cyan}{} &\Mt{green!60!white}{} \Mt{green!60!white}{} \Mt{green!60!white}{}  &\Mt{red!70!white}{} \Mt{blue!60!white}{} &\Mt{blue!60!white}{} \Mt{red!30!yellow}{} \Mt{blue!60!white}{}  &\Mt{blue!60!white}{} \Mt{red!70!white}{} &\tabincell{c}{\Mt{blue!60!white}{} \Mt{red!70!white}{} \Mt{green!60!white}{}\\\Mt{blue!60!white}{} \Mt{red!70!white}{} \Mt{green!60!white}{} \Mt{green!60!white}{}}  &\Mt{green!60!white}{} \Mt{blue!60!white}{} &\Mt{green!60!white}{} \Mt{blue!60!white}{} \Mt{green!60!white}{}  &\Mt{red!30!yellow}{} \Mt{red!70!white}{} \\
        	\hline
        	{Search}     &\Mt{cyan}{} &\Mt{red!30!yellow}{} \Mt{blue!60!white}{} \Mt{red!70!white}{}  &\Mt{green!60!white}{} \Mt{blue!60!white}{} &\Mt{red!30!yellow}{} \Mt{green!60!white}{} \Mt{green!60!white}{}  &\Mt{blue!60!white}{} \Mt{red!70!white}{} &\tabincell{c}{\Mt{red!30!yellow}{} \Mt{green!60!white}{} \Mt{blue!60!white}{}\\\Mt{red!70!white}{} \Mt{blue!60!white}{} \Mt{red!30!yellow}{} \Mt{green!60!white}{}}  &\Mt{blue!60!white}{} \Mt{blue!60!white}{} &\Mt{blue!60!white}{} \Mt{blue!60!white}{} \Mt{green!60!white}{}  &\Mt{red!70!white}{} \Mt{red!70!white}{} \\
        	
        	\hline
        \end{tabular}
	}
	\vspace{-10pt}
\end{table*}

\subsection{Asymmetric Search for Adaptive Representation}
\label{asymmetric}

Besides the fusion module, the other crucial key for vision-language tracking is how to construct the basic modeling structure. The simplest strategy is to inherit a symmetric Siamese network from vision-based tracking algorithms (\eg,~\cite{Siamfc,Siamrpn++}), as in current VL trackers~\cite{SNLT}. However, the performance gap remains when directly following the framework (see Fig.~\ref{fig:sota_comp}), which is mostly blamed on neglecting the different intrinsic nature between VL-based multimodal and vision-only single modality. To remedy this, we propose an asymmetric searching strategy (dubbed {ASearch}) to learn an adaptive modeling structure for pairing with ModaMixer.

\begin{algorithm}
	\renewcommand{\algorithmicrequire}{\textbf{Require:}}
	\begin{algorithmic}[1]
    	\STATE \textbf{Input}: 
                \emph{Network $\mathcal{N}$ with parameters $\theta$, search space $\mathcal{A}$, \\max iteration ${T}$, random sampling function $\Gamma$, \\
                Train dataset: $\mathcal{D}_{train}=\{\mathcal{X}_{n},y_{n}\}^{N}_{n=1}$, $\mathcal{X}_n=\{x^{v}_n,x^{l}_n (optional)\}$, Val dataset: $\mathcal{D}_{val}={\{\mathcal{X}_{m},y_{m}\}}^{M}_{m=1}$,
    	\\For videos without language annotation: $x^l=$``\textbf{0-tensor}'' or ``\textbf{template}'' or ``\textbf{attribute}''.}\\ \; \\
    	\STATE \textbf{Initialization}: \emph{Initialize the network parameters $\theta_{\mathcal{N}}$.}\\
    	\FOR{$i = 1$ : $T$} \label{for}
    	    \FOR{$n = 1$ : $N$}
                \IF{language annotation exists}
        	    \STATE $f^l_n=\texttt{BERT}(x^l_n)$;\\
        	\ELSIF{$x^l_n=$``\textbf{\textit{0-tensor}}''}
        	      \STATE $f^l_n=\texttt{zeros\_like}[\texttt{BERT}(x^l_n)]$;\\ \hfill {/* Default setting w/o sentence annotation */}\\
           	\ELSIF{$x^l_n=$``\textbf{\textit{template}}''}
                    \STATE $f^l_n=\texttt{ROI}(x^v_n)$;\\ \hfill {/* Robust setting w/o sentence annotation */}\\
                \ELSIF{$x^l_n=$``\textbf{\textit{attribute}}''}
                    \STATE $f^l_n=\texttt{BERT}(``none,object,none,none'')$;\\ \hfill {/* Default attribute word annotation */}\\
        	\ENDIF
        	\STATE $a=\Gamma(\mathcal{A})$, ${p}_{n}=\mathcal{N}(x^v_n,f^l_n;{a})$;\\
            \ENDFOR
    	    \STATE Update network $\theta_{\mathcal{N}}$ with gradient descent:\\
    	    $\theta_{\mathcal{N}}^{a} \leftarrow \theta_{\mathcal{N}}^{a}-\alpha \partial \frac{1}{N} \sum_{n=1}^{N}\mathcal{L}({p}_{n},y_{n};\theta_{\mathcal{N}}^{a})/\partial \theta_{\mathcal{N}}^{a}$;\\
    	\ENDFOR
    	\STATE ${a}_{best}=\texttt{EvolArchSearch}(\mathcal{A},\mathcal{D}_{val}; \theta_{\mathcal{N}})$; \hfill /* \cite{spos} */\\
     \STATE \textbf{Output}: \emph{searched optimal subnet ${a}_{best}$.\\}
 
	\end{algorithmic}  
	\caption{Asymmetric Searching Strategy (ASearch):\\Pretrain Stage.}
    \label{alg:pipeline-1}
\end{algorithm}

\begin{algorithm}
	\renewcommand{\algorithmicrequire}{\textbf{Require:}}
	\begin{algorithmic}[1]
    	\STATE \textbf{Input}: 
                \emph{Network $\mathcal{N}$ with parameters $\theta$, searched optimal subnet ${a}_{best}$, Train dataset: $\mathcal{D}_{train}=\{\mathcal{X}_{n},y_{n}\}^{N}_{n=1}$, $\mathcal{X}_n=\{x^{v}_n,x^{l}_n (optional)\}$,.}\\ \; \\
    	
    	\STATE \textbf{Initialization}: \emph{Initialize the network parameters $\theta_{\mathcal{N}}$.}\\
    	\WHILE{\emph{not converged}}
    	    \FOR{$n = 1$ : $N$}
        	    \STATE \textbf{line 5\;-\;13} in Alg.~\ref{alg:pipeline-1};\\
        	    \STATE ${p}_{n}=\mathcal{N}(x^v_n,f^l_n;{a}_{best})$;\\
    	    \ENDFOR
    	    \STATE Update network $\theta_{\mathcal{N}}$ with gradient descent:\\
    	    $\theta_{\mathcal{N}}^{a_{best}} \leftarrow \theta_{\mathcal{N}}^{a_{best}}-\alpha \partial \frac{1}{N} \sum_{n=1}^{N}\mathcal{L}({p}_{n},y_{n};\theta_{\mathcal{N}}^{a_{best}})/\partial \theta_{\mathcal{N}}^{a_{best}}$;\\
    	\ENDWHILE
        \STATE \textbf{Output}: \emph{network parameters $\theta_{\mathcal{N}}$, ${a}_{best}$.\\}
	\end{algorithmic}  
	\caption{Asymmetric Searching Strategy (ASearch):\\Retrain Stage.}
    \label{alg:pipeline-2}
\end{algorithm}

The spirits of network search are originated from the field of Neural Architecture Search (NAS). We adopt a popular NAS model, in particular the single-path one-shot method SPOS~\cite{spos}, for searching the optimal structure of our purpose. Although SPOS has been utilized for tracking~\cite{yan2021lighttrack}, our work significantly differs from it in two aspects: \textbf{1)} Our ASearch is tailored for constructing an {{asymmetric}} two-stream network for {{multimodal}} tracking, while~\cite{yan2021lighttrack} is designed to find a {{symmetric}} Siamese network for vision-only {{single-modality}} tracking. Besides, we search layers both in the backbone network and the post-processing $\operatorname{Block_{ASearch}}$ of ModaMixer (see Eq.~\ref{eq:modamixer}); \textbf{2)} ASearch reuses the pre-trained supernet from SPOS, which avoids burdensome re-training on ImageNet~\cite{imagenet} (both for supernet and found subnet) and thus reduces the time complexity of our search pipeline to $1/64$ of that in LightTrack~\cite{yan2021lighttrack} ({0.625 RTX-2080Ti GPU days \emph{v.s.} 40 V100 GPU days}). More details and comparison of our ASearch and~\cite{yan2021lighttrack} are presented in Sec.~\ref{futher_ana:b}.

\begin{table*}[t]
   \tabcolsep 2pt
   \caption{Configurations of the asymmetrical architecture learned by ASearch.}
   \label{tab:configuration}
   \centering 
  \renewcommand\arraystretch{1.05}
   \resizebox{0.99\textwidth}{!}{
   \begin{tabular}{c|c|c|c|c|c|c|c|c|c|c}
      \toprule
      &Stem &\multicolumn{2}{c|}{Stage1} &\multicolumn{2}{c|}{Stage2} &\multicolumn{2}{c|}{Stage3} &\multicolumn{2}{c|}{Stage4} &Output\\
    \midrule
    
    Layer Name &\tabincell{c}{Convolution\\Block} &\tabincell{c}{Block$_\mathrm{ASearch}$\\$\times3$} &\tabincell{c}{ModaMixer} &\tabincell{c}{Block$_\mathrm{ASearch}$\\$\times3$} &\tabincell{c}{ModaMixer} &\tabincell{c}{Block$_\mathrm{ASearch}$\\$\times7$} &\tabincell{c}{ModaMixer} &\tabincell{c}{Block$_\mathrm{ASearch}$\\$\times3$} &\tabincell{c}{ModaMixer} &\tabincell{c}{Convolution\\Block} \\
    \specialrule{0em}{1pt}{1pt}
    \midrule
    \specialrule{0em}{1pt}{1pt}
    
    Parameter &\tabincell{c}{$P_{in}=2$\\$C_{in}=16$} &\tabincell{c}{$P_1=2$\\$C_1=64$}
    & \tabincell{c}{$P=1$\\$C=64$} &\tabincell{c}{$P_2=2$\\$C_2=160$}
    & \tabincell{c}{$P=1$\\$C=160$} &\tabincell{c}{$P_3=1$\\$C_3=320$}
    & \tabincell{c}{$P=1$\\$C=320$} &\tabincell{c}{$P_4=1$\\$C_4=640$}
    & \tabincell{c}{$P=1$\\$C=640$} &\tabincell{c}{$P_{out}=1$\\${C}_{out}=256$} \\
      \specialrule{0em}{1pt}{1pt}
      \midrule
      \specialrule{0em}{1pt}{1pt}
    
    Output Size & \scalebox{1.3}{$\frac{H}{2}\times \frac{W}{2}$} &\multicolumn{2}{c|}{\scalebox{1.3}{$\frac{H}{4}\times \frac{W}{4}$}} &\multicolumn{2}{c|}{\scalebox{1.3}{$\frac{H}{8}\times \frac{W}{8}$}} &\multicolumn{2}{c|}{\scalebox{1.3}{$\frac{H}{8}\times \frac{W}{8}$}} &\multicolumn{2}{c|}{\scalebox{1.3}{$\frac{H}{8}\times \frac{W}{8}$}} &  \scalebox{1.3}{$\frac{H}{8}\times \frac{W}{8}$}\\

      \bottomrule
   \end{tabular}}
\end{table*}

The search space and search strategy of ASearch are kept consistent with the original SPOS~\cite{spos}. In particular, the search pipeline is formulated as
\begin{align}
W_\mathcal{A} &= \mathop{\argmin}_{W} \; \mathbb{E}_{a \sim \Gamma(\mathcal{A})} \left[ \mathcal{L}_\text{train}(\mathcal{N}(a, W(a))) \right], \label{eq:supernet} \\
a^* &= \mathop{\argmax}_{a\in \mathcal{A}} \; \text{SUC}_\text{val} \left( \mathcal{N} (a, W_\mathcal{A}(a)) \right), \label{eq:subnet}
\end{align}
where $\mathcal{A}$ represents the architecture search space of the network $\mathcal{N}$, $a$ is a sample from $\mathcal{A}$, and $W$ denotes the corresponding network weights. Notably, the network $\mathcal{N}$ includes three components $\mathcal{N}=\{\varphi_{t}, \varphi_{s}, \varphi_{m}\}$, where each indicates backbone for the template branch $\varphi_{t}$, backbone for the search branch $\varphi_{s}$ and layers in the ModaMixer $\varphi_{m}$. The whole pipeline consists of training supernet on tracking datasets via random sampling $\Gamma$ from search space $\mathcal{A}$ (Eq.~\ref{eq:supernet}) and finding the optimal subnet via evolutionary algorithms (Eq.~\ref{eq:subnet}). The SUC (success score) on validation data is used as rewards for evolutionary algorithms. Table~\ref{tab:framework} demonstrates the searched asymmetric networks in our VL tracking.

The pipeline of ASearch consists of two stages. {The first stage pretrains to search an optimal architecture and the second one retrains it for our VL tracking}, as shown in Alg.~\ref{alg:pipeline-1} and Alg.~\ref{alg:pipeline-2}. The pretraining stage (Alg.~\ref{alg:pipeline-1}) contains four steps: \textbf{1)} ASearch first initializes the network parameter $\theta_{\mathcal{N}}$ of $\mathcal{N}$. Concretely, $\varphi_{t}$ and $\varphi_{s}$ reuse the pretrained supernet of SPOS~\cite{spos}, while $\varphi_{m}$ copies the weight of the last layer in $\varphi_{t}$ and $\varphi_{s}$. This reduces the tedious training on ImageNet~\cite{imagenet} and enables quick reproducibility of our work; \textbf{2)} The language model~\cite{bert} processes the language annotation ${x}^{l}$ to get corresponding representation $f^{l}$. If the language annotation is not provided, three different strategies are designed to handle these cases (\ie, ``{0-tensor}'', ``{template}'' or ``{attribute}'', illustrated in Sec.~\ref{futher_ana:a}); \textbf{3)} Then, $\theta_{\mathcal{N}}$ is trained for $T$ iterations. For each iteration, a subnet $a$ is randomly sampled from search space $\mathcal{A}$ by the function $\Gamma$ and outputs the predictions $p$. The corresponding parameters of $a$ would be updated by gradient descent; \textbf{4)} After pretraining, Evolutionary Architecture Search~\cite{spos} is performed to find the optimal subnet ${a}_{best}$. The rewarding for the evolutionary search is the SUC (success score) on the validation data $\mathcal{D}_{val}$. 

The retraining stage (Alg.~\ref{alg:pipeline-2}) is to optimize the searched subnet ${a}_{best}$ following the training pipeline of baseline trackers~\cite{SiamCAR,TransT,ostrack}. Table~\ref{tab:configuration} details the configurations of the searched asymmetric architecture.

\begin{figure}[!t]
\centering
\includegraphics[width=\linewidth]{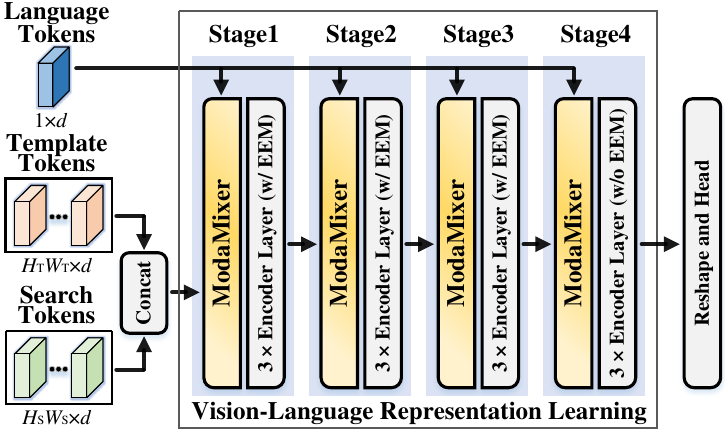}
\vspace{-11pt}
\caption{
Framework of OSTrack~\cite{ostrack} with our proposed ModaMixer. ``EEM'' denotes the Early Elimination Module in~\cite{ostrack}.
}
\label{fig:ostrack-modamixer}
\end{figure}

\subsection{Tracking Framework}
\label{method:framework}

With the proposed attribute annotation strategy, ModaMixer and the searched asymmetric networks, we construct a new vision-language tracking framework, as shown in Fig.~\ref{fig:framework} and Table~\ref{tab:framework}. Our framework is matching-based tracking. Both template and search backbone networks contain 4 stages with the maximum stride of 8, the chosen blocks of each stage are denoted with different colors in Table~\ref{tab:framework}. ModaMixer is integrated into each stage of the template and search networks to learn informative mixed representation. It is worth noting that, the asymmetry is revealed in not only the design of backbone networks, but also the ModaMixer. Each ModaMixer shares the same meta-structure as in Fig.~\ref{fig:modamixer}, but comprises different post-processing layers $\operatorname{Block_{ASearch}}$ to allow adaption to different semantic levels (\ie, network depth) and input signals (\ie, template and search, pure-vision and mixed feature in each ModaMixer). With the learned unified-adaptive VL representations from the template and search branches, we perform feature matching and target localization, the same as in the Siamese baselines (\ie, SiamCAR~\cite{SiamCAR} and TransT~\cite{TransT}). Notably, for the one-stream baseline OSTrack~\cite{ostrack}, we apply the proposed ModaMixer into each backbone stage to enhance the joint representation learning with linguistic semantics, as shown in Fig.~\ref{fig:ostrack-modamixer}. The other modules and settings follow the baseline without any modifications.

\subsection{A Theoretical Explanation}
\label{theoretical}

This section presents a theoretical explanation of our method, following the analysis in \cite{Prove}. Based on the Empirical Risk Minimization (ERM) principle~\cite{ERM}, the objective of representation learning is to find better network parameters $\theta$ by minimizing the empirical risk, \textit{i.e.},
\begin{equation}
    \min \;\;\hat{r}\left(\theta_{\mathcal{M}}\right)\triangleq \frac{1}{n}\sum_{i=1}^{n}\mathcal{L}\left(\mathcal{X}_i,y_{i};\theta_{\mathcal{M}}\right) \;\;\;\;\;
    \text{s.t.}  \;\; \theta_{\mathcal{M}}\in\mathcal{F}. 
\end{equation}
where $n$ indicates sample number, $\mathcal{L}$ denotes loss function, $\mathcal{M}$ represents the modality set with size of $|\mathcal{M}|$, $\mathcal{X}_i = \{x_{i}^{1}, x_{i}^{2} ... x_{i}^{|\mathcal{M}|}\}$ is the input multimodal signal, $y_{i}$ is training label, $\theta_{\mathcal{M}}$ is the searched network parameters with $\mathcal{M}$, and $\mathcal{F}$ denotes optimization space of $\theta$. Given the empirical risk $\hat{r}\left(\theta_{\mathcal{M}}\right)$, its corresponding population risk is defined as,
\begin{eqnarray}
r\left(\theta_{\mathcal{M}}\right)=\mathbb{E}_{(\mathcal{X}_i,y_{i})\sim\mathcal{D}_{train}} \left[\hat{r}\left(\theta_{\mathcal{M}}\right)\right]    
\end{eqnarray}
Following \cite{risk1,risk2,Prove}, the population risk is adopted to measure the learning quality. Then the {latent representation quality}~\cite{Prove} is defined as,
\begin{align}
   \eta(\theta) = \inf_{\theta\in \mathcal{F}}\left[ r\left(\theta\right)-r(\theta^{*})\right]
\end{align}
where $*$ represents the theoretical optimal case, $\inf$ indicates the best achievable population risk. With the empirical Rademacher complexity $\mathfrak{R}$~\cite{Rademacher}, we restate the conclusion in \cite{Prove} with our definition.

\begin{theorem}[\cite{Prove}]\label{thm:mn-modality}
Let $\mathcal{S}$ be another modality set with size of $|\mathcal{S}|$ ($|\mathcal{M}| \; \textgreater \; |\mathcal{S}|$). Assuming we have produced the empirical risk minimizers $\hat{\theta}_{\mathcal{M}}$ and $\hat{\theta}_{\mathcal{S}}$, training with the $|\mathcal{M}|$ and $|\mathcal{S}|$ modalities separately. Then, for all $\delta\in(0,1)$, with probability at least $1-\frac{\delta}{2}$:
\begin{equation}
\begin{split}
r\left(\hat{\theta}_{\mathcal{M}}\right)- r\left(\hat{\theta}_{\mathcal{S}}\right)
&\leq \gamma_{\mathcal{D}}(\mathcal{M},\mathcal{S})+8L\mathfrak{R}_{n}(\mathcal{F}_{\mathcal{M}})\\
&+\frac{4C}{\sqrt{n}}+2C\sqrt{\frac{2 \ln (2 / \delta)}{n}}\label{mg}
\end{split}
\end{equation}
where
\begin{equation}
\begin{split}
\gamma_{\mathcal{D}}(\mathcal{M},\mathcal{S})\triangleq \eta(\hat{\theta}_{\mathcal{M}})-\eta(\hat{\theta}_{\mathcal{S}})\label{gamma}\qquad\\
\mathfrak{R}_{n}(\mathcal{F}_{\mathcal{M}})\sim \sqrt{\mathrm{Complexity}(\mathcal{F}_{\mathcal{M}})/n} 
\end{split}
\end{equation}
\end{theorem}
\noindent$\gamma_{\mathcal{D}}(\mathcal{M},\mathcal{S})$ computes the quality difference learned from multiple modalities $\mathcal{M}$ and single modality $\mathcal{S}$ with dataset $\mathcal{D}$. Theorem \ref{thm:mn-modality} defines an upper bound of the population risk training with different number of modalities, {which proves that more modalities could potentially enhance the representation quality.} Furthermore, the Rademacher complexity $\mathfrak{R}_{n}(\mathcal{F})$ is proportional to the network complexity, {which demonstrates that heterogeneous network would theoretically rise the upper bound of $r\left(\hat{\theta}_{\mathcal{M}}\right)- r\left(\hat{\theta}_{\mathcal{S}}\right)$, and also exhibits that our searched asymmetric design has larger optimization space when learning with $|\mathcal{M}|$ modalities compared to $|\mathcal{S}|$ modalities ($|\mathcal{M}|>|\mathcal{S}|$).} The proof can be referred to~\cite{Prove} for details.

\begin{table*}[!t]
  \begin{center}
  \caption{Comparison with State-of-the-arts on LaSOT~\cite{LaSOT}, LaSOT$_\mathrm{Ext}$~\cite{LaSOT_Extention}, TNL2K~\cite{TNL2K}, GOT-10k~\cite{GOT10K}, OTB99-LANG (OTB99-L)~\cite{Li} and TrackingNet~\cite{trackingnet}. \textcolor{black}{OSTrack}, \textcolor{black}{TransT} and \textcolor{black}{SiamCAR} are baselines of proposed {VLT}$_{\mathrm{OST}}$, {VLT}$_{\mathrm{TT}}$ and {VLT}$_{\mathrm{SCAR}}$, respectively. SUC, P, AO, SR denote metrics of success score, precision score, average overlap and success rate, respectively. All metrics of performance are in \% in tables unless otherwise specified.}
  \vspace{0.3em}
  \label{tab:sota}
  \newcommand{\RED}[1]{\textcolor[rgb]{1.00,0.00,0.00}{#1}}
  \newcommand{\dist}{\hspace{1pt}}
  \renewcommand\arraystretch{1.15}
  \resizebox{0.99\linewidth}{!}{
  \begin{tabular}{c|cc|cc|cc|ccc|cc|cc}
  \hline
  
  \hline
  \multirow{2}{*}{{Method}}  &\multicolumn{2}{c|}{{LaSOT}}	&\multicolumn{2}{c|}{{LaSOT$_\mathrm{Ext}$}}
  &\multicolumn{2}{c|}{{TNL2K}}	&\multicolumn{3}{c|}{{GOT-10k}} &\multicolumn{2}{c|}{{OTB99-L}}  &\multicolumn{2}{c}{{TrackingNet}} \\
  \cline{2-14}
  &{SUC}		&{P} &{SUC}		&{P}	&{SUC}  &{P}	&{AO} &{SR$_{0.5}$} &{SR$_{0.75}$} &{SUC}  &{P}  &{SUC}  &{P}\\
  \hline
   
  \hline
  ECO~\cite{ECO}	&32.4  &30.1
  &22.0  &24.0	
  &-  &-
  &31.6 &30.9 &11.1
  &-  &- 
  &55.4 &49.2\\
  SiamFC~\cite{Siamfc}	&33.6  &33.9
  &23.0  &26.9	
  &29.5  &28.6
  &34.8 &35.3 &9.8
  &58.7  &79.2 
  &57.1 &53.3\\
  C-RPN~\cite{C-RPN}	&45.5  &42.5
  &27.5  &32.0	
  &-  &-
  &- &- &-
  &-  &- 
  &66.9 &61.9\\
  SiamRPN++~\cite{Siamrpn++}	&49.6  &49.1
  &34.0  &39.6	
  &41.3  &41.2
  &51.7 &61.6 &32.5
  &63.8 &82.6 
  &73.3 &69.4\\
  \textcolor{black}{SiamCAR}~\cite{SiamCAR}	&50.7  &51.0
  &33.9  &41.0	
  &35.3  &38.4
  &56.9 &67.0 &41.5
  &68.8  &89.1 
  &65.4 &66.6\\
  ATOM~\cite{Atom}	&51.5  &50.5
  &37.6  &43.0	
  &40.1  &39.2
  &55.6 &63.4 &40.2
  &67.6  &82.4 
  &70.3 &64.8\\
  SNLT~\cite{SNLT}	&54.0  &57.6
  &26.2  &30.0	
  &27.6  &41.9
  &43.3 &50.6 &22.1
  &66.6  &80.4 
  &- &-\\
  KYS~\cite{kys}	&55.4  &-
  &-  &-	
  &44.9  &43.5
  &63.6 &75.1 &51.5
  &-  &- 
  &74.0 &68.8\\
  Ocean~\cite{Ocean}	&56.0  &56.6
  &-  &-	
  &38.4  &37.7
  &61.1 &72.1 &47.3
  &68.0  &92.1 
  &- &-\\
  AutoMatch~\cite{AutoMatch}	&58.3  &59.9
  &37.6  &43.0	
  &47.2  &43.5
  &65.2 &76.6 &54.3
  &71.6  &93.2 
  &76.0 &72.6\\
  PrDiMP~\cite{prdimp}	&59.8  &60.8
  &-  &-	
  &47.0  &45.9
  &63.4 &73.8 &54.3
  &69.5  &89.5 
  &75.8 &70.4\\
  SiamRCNN~\cite{siamrcnn}	&64.8  &68.4
  &-  &-	
  &52.3  &52.8
  &64.9 &72.8 &59.7
  &70.0  &89.4 
  &81.2 &80.0\\



  \hline
  
  TrDiMP~\cite{TrDiMP}	&63.9  &66.3
  &-  &-	
  &-  &-
  &67.1 &77.7 &58.3
  &70.5  &92.5 
  &78.4 &73.1\\
  \textcolor{black}{TransT}~\cite{TransT}	&64.9  &69.0
  &44.8  &52.5	
  &50.7  &51.7
  &67.1 &76.8 &60.9
  &70.8  &91.2 
  &81.4 &80.3\\
  STARK-50~\cite{stark}	&66.4  &71.2
  &47.8  &55.1
  &-  &-
  &68.0 &77.7 &62.3
  &69.6  &91.4 
  &81.3 &-\\
  SwinTrack~\cite{swintrack}	&67.2  &70.8
  &47.6  &53.9	
  &53.0  &53.2
  &71.3 &81.9 &64.5
  &-  &- 
  &81.3 &78.4\\
  \textcolor{black}{OSTrack-256}~\cite{ostrack}	&69.1  &75.2
  &47.4  &53.3	
  &54.3  &-
  &71.0 &80.4 &68.2
  &76.5  &92.7 
  &83.1 &82.0\\
  MixFormer-22k~\cite{mixformer}	&69.2  &74.7
  &-  &-	
  &-  &-
  &72.6 &82.2 &68.8
  &71.0  &93.1 
  &83.1 &81.6\\

  \hline

  

  \cellcolor{lightgray!40}{{VLT$_{\mathbf{\mathrm{SCAR}}}$} (Ours)}	&\cellcolor{lightgray!40}{\textbf{68.1}}  &\cellcolor{lightgray!40}{\textbf{73.6}}
  &\cellcolor{lightgray!40}{\textbf{47.4}}  &\cellcolor{lightgray!40}{\textbf{55.1}}	
  &\cellcolor{lightgray!40}{\textbf{52.9}}  &\cellcolor{lightgray!40}{\textbf{53.9}}
  &\cellcolor{lightgray!40}{\textbf{65.5}} &\cellcolor{lightgray!40}{\textbf{78.1}} &\cellcolor{lightgray!40}{\textbf{55.8}}
  &\cellcolor{lightgray!40}{\textbf{75.4}}  &\cellcolor{lightgray!40}{\textbf{90.9}}
  &\cellcolor{lightgray!40}{\textbf{80.4}}  &\cellcolor{lightgray!40}{\textbf{77.3}}\\

  

  \cellcolor{lightgray!40}{{{VLT$_{\mathbf{\mathrm{TT}}}$} (Ours)}}	&\cellcolor{lightgray!40}{{\textbf{68.8}}} &\cellcolor{lightgray!40}{{\textbf{75.4}}}
  &\cellcolor{lightgray!40}{{\textbf{49.7}}}  &\cellcolor{lightgray!40}{{\textbf{57.4}}}	
  &\cellcolor{lightgray!40}{{\textbf{55.0}}}  &\cellcolor{lightgray!40}{{\textbf{54.2}}}
  &\cellcolor{lightgray!40}{{\textbf{71.4}}} &\cellcolor{lightgray!40}{{\textbf{83.8}}} &\cellcolor{lightgray!40}{{\textbf{64.1}}}
  &\cellcolor{lightgray!40}{{\textbf{77.2}}}  &\cellcolor{lightgray!40}{{\textbf{93.9}}} 
  &\cellcolor{lightgray!40}{\textbf{83.1}}  &\cellcolor{lightgray!40}{\textbf{81.3}}\\

  \cellcolor{lightgray!40}{{{VLT$_{\mathbf{\mathrm{OST}}}$} (Ours)}}	&\cellcolor{lightgray!40}{{\textbf{70.5}}} &\cellcolor{lightgray!40}{{\textbf{77.0}}}
  &\cellcolor{lightgray!40}{{\textbf{48.6}}}  &\cellcolor{lightgray!40}{{\textbf{54.8}}}	
  &\cellcolor{lightgray!40}{{\textbf{55.7}}}  &\cellcolor{lightgray!40}{{\textbf{54.5}}}
  &\cellcolor{lightgray!40}{{\textbf{72.7}}} &\cellcolor{lightgray!40}{{\textbf{83.9}}} &\cellcolor{lightgray!40}{{\textbf{68.1}}}
  &\cellcolor{lightgray!40}{{\textbf{77.7}}}  &\cellcolor{lightgray!40}{{\textbf{95.0}}} 
  &\cellcolor{lightgray!40}{\textbf{84.3}}  &\cellcolor{lightgray!40}{\textbf{83.1}}\\
   
  \hline
  
  \hline
  \end{tabular}}
  \end{center}

  \end{table*}

\section{Experiment}

\subsection{Implementation Details}
\label{implementation_details}
We apply our method to CNN-based SiamCAR~\cite{SiamCAR}, CNN-Transformer-hybrid TransT~\cite{TransT} and Transformer-based OSTrack~\cite{ostrack} (dubbed {VLT$_{\mathrm{SCAR}}$}, {VLT$_{\mathrm{TT}}$} and {VLT$_{\mathrm{OST}}$}, respectively). The other modules are inherited from the baseline tracker without any modifications.  

\textbf{Datasets for VLT.} To demonstrate the effectiveness and generality of our VL database, we respectively train and infer our tracker with official sentence annotation and the proposed attribute words. OTB99-Language~\cite{Li} is used only for evaluation). Notably, GOT-10k~\cite{GOT10K} provides descriptions for object/motion/major/root class, \eg, ``dove, walking, bird, animal'', in each video. We concatenate these words to obtain a pseudo sentence description, and the major/root class is inherited in our attribute annotation. For the datasets without language description (\eg, COCO~\cite{COCO}), the language representation is set to 0-tensor, visual pooling feature, or ``category, object, none, none''. We discuss the different settings in Sec.~\ref{futher_ana:a}.

\textbf{Searching for VLT.} The proposed ASearch aims to find a more flexible modeling structure for vision-language tracking (VLT). Taking VLT$_{\mathrm{SCAR}}$ as example, the supernet from SPOS~\cite{spos} is used as feature extractor to replace the ResNet~\cite{resnet} in SiamCAR. We train the trackers with supernet using training splits of COCO~\cite{COCO}, ImageNet-VID~\cite{imagenet}, ImageNet-DET~\cite{imagenet}, YouTube-BB~\cite{youtube}, GOT-10k~\cite{GOT10K}, LaSOT~\cite{LaSOT} and TNL2K~\cite{TNL2K} for 5 epochs, where each epoch contains $1.2\times10^{6}$ template-search pairs. Once finishing supernet training, evolutionary algorithms as in SPOS~\cite{spos} are applied to search for optimal subnet and finally obtain VLT$_{\mathrm{SCAR}}$. The whole search pipeline consumes 15 hours on a single RTX-2080Ti GPU. The search process of VLT$_{\mathrm{TT}}$ is similar to VLT$_{\mathrm{SCAR}}$. Notably, for one-stream VLT$_{\mathrm{OST}}$, we only search the $\operatorname{Block_{ASearch}}$ of ModaMixer in each backbone stage, as shown in Fig.~\ref{fig:ostrack-modamixer}.

\begin{table}[!t]
  \caption{Ablation on ModaMixer and asymmetric searching strategy (ASearch).}
  \label{tab:ablation}
  \begin{center}
  \newcommand{\dist}{\hspace{3pt}}
  \newcommand{\best}[1]{\textbf{\textcolor{red}{#1}}}
  \newcommand{\scnd}[1]{\textbf{\textcolor{blue}{#1}}}
  \renewcommand\arraystretch{1.2}
  
  \resizebox{1\linewidth}{!}{
  \begin{tabular}{@{}c|c|c|c|cc|cc@{}}
  \hline
  \multirow{2}{*}{\#}& \multirow{2}{*}{{Method}}& Moda-   &  \multirow{2}{*}{{ASearch}} &  \multicolumn{2}{c|}{{LaSOT}} &\multicolumn{2}{c}{{TNL2K}}  \\
  \cline{5-8}
  & & Mixer&  &{SUC}  &{P} &{SUC} &{P} \\
  \hline
  \ding{172}&Baseline &-   &-   &50.7 &51.0 &{35.3} &38.4 \\
  
  \ding{173}&{VLT}$_{\mathrm{SCAR}}$&$\surd$ &-           &57.6	 &61.1 &{41.5} &43.2 \\
  \ding{174}&{VLT}$_{\mathrm{SCAR}}$  &- &$\surd$  &52.1	 &50.6 &{40.7}&40.2 \\
  
  \ding{175}&{VLT}$_{\mathrm{SCAR}}$ &$\surd$ &$\surd$	&{65.2} &{69.1} &{48.3} &{46.6} \\

  \hline
  \end{tabular}}
  \end{center}
  \end{table}

\textbf{Optimizing for VLT.} The training protocol of VLT$_{\mathrm{SCAR}}$, VLT$_{\mathrm{TT}}$ and VLT$_{\mathrm{OST}}$ follow the corresponding baselines (\ie, SiamCAR~\cite{SiamCAR}, TransT~\cite{TransT} and OSTrack~\cite{ostrack}). Notably, for each epoch, half training pairs come from datasets without language annotations (\eg, COCO~\cite{COCO}, ImageNet-DET~\cite{imagenet}, YouTube-BB~\cite{youtube}), which aims to enhance the robustness for tracking with the only vision modality (discussed in Sec.~\ref{futher_ana:a}). We only introduce the contrastive loss of ModaMixer in the first backbone stage to facilitate vision-language alignment (analyzed in Sec.~\ref{futher_ana:b}).

\subsection{Comparison with State-of-the-arts}
\label{sota_comparison}
Table~\ref{tab:sota} presents the results and comparisons of our trackers with other SOTAs on LaSOT~\cite{LaSOT}, LaSOT$_{\mathrm{Ext}}$~\cite{LaSOT_Extention}, TNL2K~\cite{TNL2K}, GOT-10K~\cite{GOT10K} , OTB99-LANG~\cite{Li} and TrackingNet~\cite{trackingnet}. The proposed VLT$_{\mathrm{SCAR}}$, VLT$_{\mathrm{TT}}$ and VLT$_{\mathrm{OST}}$ run at 43/35/31 FPS on a single RTX-2080Ti GPU, respectively, while the baseline trackers SiamCAR~\cite{SiamCAR}/TransT~\cite{TransT}/OSTrack~\cite{ostrack} runs at 52/32/33 FPS. Interestingly, our VLT$_{\mathrm{TT}}$ outperforms the baseline TransT in terms of both accuracy and speed.

Particularly, compared with SiamCAR~\cite{SiamCAR}, our VLT$_{\mathrm{SCAR}}$ achieves considerable SUC gains of 17.4\%/13.5\%/17.6\% on LaSOT/LaSOT$_{\mathrm{Ext}}$/TNL2K, respectively, which demonstrates the effectiveness of the proposed VL tracker. Notably, our VLT$_{\mathrm{SCAR}}$ outperforms the current best VL tracker SNLT~\cite{SNLT} for 14.1\%/21.2\% on LaSOT/LaSOT$_{\mathrm{Ext}}$, showing that the unified-adaptive vision-language representation is more robust for VL tracking and the proposed framework is superior than simply use language description as a prompt. The advancement of our method is preserved across different benchmarks. What surprises us more is that the CNN-based VLT$_{\mathrm{SCAR}}$ is competitive and even better than recent vision Transformer-based approaches. For example, VLT$_{\mathrm{SCAR}}$ outperforms TransT~\cite{TransT} on LaSOT and meanwhile runs faster (43 FPS \emph{v.s.} 32 FPS) and requires less training pairs ($2.4\times 10^{7}$ \emph{v.s.} $3.8 \times 10^{7}$). By applying our method to Transformer-based TransT and OSTrack, the new trackers VLT$_{\mathrm{TT}}$/VLT$_{\mathrm{OST}}$ improve the baselines to 68.8\%/70.5\% in SUC with 3.9\%/1.4\% gains on LaSOT while bringing little computation cost, showing its generality.

\subsection{Component-wise Ablation}
We analyze the influence of each component in our method to show the effectiveness and rationality of the proposed ModaMixer and ASearch. The results are presented in Table~\ref{tab:ablation}. By directly applying the ModaMixer on the baseline SiamCAR~\cite{SiamCAR} (``ResNet50+ModaMixer''), it obtains SUC gains of $6.9\%$ on LaSOT (\ding{173} \emph{v.s.} \ding{172}). This verifies that the unified VL representation effectively improves the tracking robustness. One interesting observation is that ASearch improves the vision-only baseline SiamCAR~\cite{SiamCAR} for only $1.4\%$ percents on LaSOT (\ding{174} \emph{v.s.} \ding{172}), but when equipping with ModaMixer, it surprisingly further brings $7.6\%$ SUC gains (\ding{175} \emph{v.s.} \ding{173}), which shows the complementarity of multimodal representation learning (ModaMixer) and the proposed ASearch.

\subsection{Further Analysis}
\label{futher_ana}

\subsubsection{Data and Annotation}
\label{futher_ana:a}

\textbf{Dealing with Data without Language Annotation during Training.} As mentioned above, language annotations are not provided in several training datasets (\eg, COCO~\cite{COCO}, ImageNet-DET~\cite{imagenet}, YouTube-BB~\cite{youtube}). We design three strategies to replace the missed language representation (two for {sentence} annotation and one for {attribute word} description). Specifically, when using sentence as the language annotation, it is hard to prepare a specialized sentence to describe each training pair (generating the pseudo sentence description with an image-caption model will be discussed later). One simple way is to use ``0-tensor'' as language embedding, and the other is to replace the language embedding with visual features by pooling template features in the bounding box. As shown in Table~\ref{tab:sota_ours}, the two strategies (\textit{i.e.,} VLT$^{\:t}_{\mathbf{\mathrm{SCAR}}}$ and VLT$^{\:t}_{\mathbf{\mathrm{TT}}}$) show similar performances, while the one with visual feature is slightly better in average. As for the attribute version, we annotate the data with ``category, object, none, none'' for the datasets with category name (\textit{e.g.} COCO~\cite{COCO}), and ``none, object, none, none'' for other datasets without category information (represented by ``attribute'' in Alg.~\ref{alg:pipeline-1}). Table~\ref{tab:sota_ours} shows that the attribute versions (\textit{i.e.,} VLT$^{\:attr}_{\mathbf{\mathrm{SCAR}}}$ and VLT$^{\:attr}_{\mathbf{\mathrm{TT}}}$) achieve better performance compared with the ones using sentence description, proving our claim that attribute words are more general and effective to learn VL representation for tracking.

\begin{table}[!t]
  \begin{center}
  \caption{Comparison of our trackers with different data settings. $^0$ and $^t$ denote the settings of ``0-tensor'' and ``template'' for data without {sentence} annotation, and $^{attr}$ represents the ``attribute'' setting which uses {attribute words} as the language description. $\ddagger$ denotes training with contrastive loss~\cite{clip}. All metrics of performance are in \% in tables unless otherwise specified.}
  \label{tab:sota_ours}
  \newcommand{\RED}[1]{\textcolor[rgb]{1.00,0.00,0.00}{#1}}
  \newcommand{\dist}{\hspace{1pt}}
  \renewcommand\arraystretch{1.15}
  \resizebox{1\linewidth}{!}{
  \begin{tabular}{c|cc|cc|ccc}
  
  \hline
  \multirow{2}{*}{{Method}}  &\multicolumn{2}{c|}{{LaSOT}}
  &\multicolumn{2}{c|}{{TNL2K}}	&\multicolumn{3}{c}{{GOT-10k}}  \\
  \cline{2-8}
  &{SUC}		&{P}	&{SUC}  &{P}	&{AO} &{SR$_{0.5}$} &{SR$_{0.75}$} \\
  \hline
   

  
  {{VLT$^{\:0}_{\mathbf{\mathrm{SCAR}}}$} }	&{{65.2}}  &{{69.1}}	
  &{{48.3}}  &{{46.6}}
  &{{61.4}} &{{72.4}} &{{52.3}}\\
  
  {{VLT$^{\:t}_{\mathbf{\mathrm{SCAR}}}$} }	&{{63.9}}  &{{67.9}}	
  &{{49.8}}  &{{51.1}}
  &{{61.0}} &{{70.8}} &{{52.2}} \\

  {{VLT$^{\:attr}_{\mathbf{\mathrm{SCAR}}}\ddagger$} }	&{{68.1}}  &{{73.6}}
  &{{52.9}}  &{{53.9}}
  &{{65.5}} &{{78.1}} &{{55.8}}\\
   
  {{{VLT$^{\:0}_{\mathbf{\mathrm{TT}}}$} }}	&{{{66.3}}} &{{{70.5}}}
  &{{{52.2}}}  &{{{52.1}}}
  &{{{68.4}}} &{{{81.5}}} &{{{62.4}}}\\
  
  {{{VLT$^{\:t}_{\mathbf{\mathrm{TT}}}$} }}	&{{{67.3}}} &{{{72.1}}}
  &{{{53.1}}}  &{{{53.3}}}
  &{{{69.4}}} &{{{81.1}}} &{{{64.5}}}\\

  {{{VLT$^{\:attr}_{\mathbf{\mathrm{TT}}}\ddagger$} }}	&{{{68.8}}} &{{{75.4}}}	
  &{{{55.0}}}  &{{{54.2}}}
  &{{{71.4}}} &{{{83.8}}} &{{{64.1}}} \\

  {{{VLT$^{\:attr}_{\mathbf{\mathrm{OST}}}\ddagger$} }}	&{{{70.5}}} &{{{77.0}}}
  &{{{55.7}}}  &{{{54.5}}}
  &{{{72.7}}} &{{{83.9}}} &{{{68.1}}}\\
   
  
  \hline
  \end{tabular}}
  \end{center}

  \end{table}

\textbf{Dealing with Data without Language Annotation during Inference.} VL trackers require language annotation on
the first frame of a video for inference. One may wonder that what if there is no language description? Table~\ref{tab:language} presents the results based on VLT$_{\mathbf{\mathrm{SCAR}}}$ and VLT$_{\mathbf{\mathrm{TT}}}$ with four different settings (\ie, ``0-tensor'', ``template'', ``attribute'' and ``Pse. language''). The results show that, without language description, tracking performances heavily degrade compared with Table~\ref{tab:sota_ours} (\textit{e.g.,} $63.9\% \rightarrow 53.4\%$, $67.3\% \rightarrow 61.0\%$ SUC on LaSOT of VLT$^{\:t}_{\mathbf{\mathrm{SCAR}}}$ and VLT$^{\:t}_{\mathbf{\mathrm{TT}}}$, respectively), verifying that the high-level semantics in language do help in improving robustness. Even though, the performances of VLT$_{\mathbf{\mathrm{SCAR}}}$ are still better than the vision-only baseline. Besides, the attribute-based version (\ie, ``attribute'') shows better performance compared with the sentence-based ones (\ie, ``0-tensor'' and ``template''), demonstrating the effectiveness of our proposed annotation strategy. Notably, when using the generated sentence description by a recent advanced image-caption method~\cite{clip}, the results are not good as expected (\textit{e.g.,} $51.6\%$ of VLT$^{\:t}_{\mathbf{\mathrm{SCAR}}}$ and $59.7\%$ of VLT$^{\:t}_{\mathbf{\mathrm{TT}}}$), indicating that it is still challenging to generate accurate caption in real-world cases and noisy caption with different sentence description style would bring negative effects to the model.

\textbf{Different Attribute Annotation during Inference.} The annotated four attribute words of each target provide high-level linguistic information from different perspectives. We further explore the influence by training and inference with different attribute words based on VLT$_{\mathbf{\mathrm{SCAR}}}^{\:attr}$ and VLT$_{\mathbf{\mathrm{TT}}}^{\:attr}$. As shown in Table~\ref{tab:attr_wonlp}, VLT$_{\mathbf{\mathrm{SCAR}}}^{\:attr}$ with the only ``root class'' attribute has already outperformed the sentence-based VLT$_{\mathbf{\mathrm{SCAR}}}^{\:t}$ for $1.6\%$ SUC of LaSOT (\ding{174} \textit{v.s.} \ding{172}). When equipping with all the four attribute words, VLT$_{\mathbf{\mathrm{SCAR}}}^{\:attr}$ (\ding{178}) obtains the best performance with $66.4\%/51.8\%$ SUC of LaSOT/TNL2K, respectively. For inference without language (\ie, using the default ``category name, object, none, none'' as the attribute annotation), VLT trained with different attribute words surpass that trained with sentences (\ding{179},\ding{177},\ding{175} \textit{v.s.} \ding{173}). Besides, VLT$_{\mathbf{\mathrm{SCAR}}}^{\:attr}$ performs more robust to the situation of missing language description compared with VLT$_{\mathbf{\mathrm{SCAR}}}^{\:t}$ (\eg, $65.5\% \rightarrow 59.0\%$ of \ding{174},\ding{175} \textit{v.s.} $63.9\% \rightarrow 53.4\%$ of \ding{172},\ding{173}). These prove our claim that the proposed attribute annotation is more general and robust for VL representation learning compared with circumscribed sentence description. One interesting observation is that the version trained with more attribute words shows higher performance compared the less ones when inference with language (\eg, \ding{178} \textit{v.s.} \ding{174}), however, the situation reverses without using the language annotation (\eg, \ding{179} \textit{v.s.} \ding{175}). This indicates that training with more specific language description would increase the dependence on linguistic modality, which also demonstrates the generality of our proposed attribute annotation instead of circumscribed sentence description.


\begin{table}[t]
   \caption{Comparisons with different settings for inference without language annotation. ``Pse. language'' denotes using the generated sentence description by an image-caption method~\cite{clip}.}
   \label{tab:language}
   \begin{center}
   \newcommand{\dist}{\hspace{1pt}}
   \renewcommand\arraystretch{1.2}
   \resizebox{1\linewidth}{!}{
   \begin{tabular}{c|c|ccc|ccc}
   \hline
   \multirow{2}{*}{{Method}} &\multirow{2}{*}{{Settings}}  &\multicolumn{3}{c|}{{LaSOT}}   &\multicolumn{3}{c}{{TNL2K}} \\
   \cline{3-8}
   & &{SUC}  &{P}$_{\mathrm{Norm}}$		&{P} &{SUC} &{P}$_{\mathrm{Norm}}$	&{P}	\\
  \hline
   \multirow{4}{*}{\tabincell{c}{{VLT$_{\mathbf{\mathrm{SCAR}}}$}}} 
   
   &{{0-tensor}} &50.8  &57.9   &52.6  &39.5 &47.1 &41.2 \\ \cline{2-8}
   &{{template}} &53.4  &60.7    &54.6  &41.1 &49.1 &42.9 \\ \cline{2-8}
   &{{attribute}} &57.1  &64.6  &60.5  &45.7 &52.2 &46.3 \\ \cline{2-8}
   &{{Pse. language}} &51.6 &58.1 &53.4  &38.8 &46.6 &40.5 \\
   
   \hline

   \multirow{4}{*}{\tabincell{c}{{VLT$_{\mathbf{\mathrm{TT}}}$}}} 
   
   &{{0-tensor}} &60.7  &71.1  &63.1  &48.2 &54.6 &46.8 \\ \cline{2-8}
   &{{template}} &61.0  &71.5  &63.4  &49.1 &55.5 &48.3 \\ \cline{2-8}
   &{{attribute}} &61.7 &73.4	&66.8	&50.6 &56.8	&51.4 \\ \cline{2-8}
   &{{Pse. language}} &59.7  &69.2 &63.0  &50.0 &56.3	&50.3 \\

   \hline
   \end{tabular}}
   \end{center}
   \end{table}

\begin{table*}[t]
  \caption{Ablation on different {attribute words} for inference. ``w/. Language'' denotes inference with language annotation. Contrastive loss~\cite{clip} is not used.}
  \label{tab:attr_wonlp}
  \begin{center}
  \newcommand{\dist}{\hspace{3pt}}
  \newcommand{\best}[1]{{\textcolor{red}{#1}}}
  \newcommand{\scnd}[1]{{\textcolor{blue}{#1}}}
  \renewcommand\arraystretch{1.2}
  
  \resizebox{0.99\linewidth}{!}{
  \begin{tabular}{c|c|cccc|c|ccc|ccc}
  \hline
  \multirow{2}{*}{\#}& \multirow{2}{*}{{Method}}& \multicolumn{4}{c|}{{\tabincell{c}{Attribute Words}}}    &  \multirow{2}{*}{{\tabincell{c}{w/.\\Language}}} &  \multicolumn{3}{c|}{{LaSOT}} &\multicolumn{3}{c}{{TNL2K}}  \\
  \cline{3-6}
  \cline{8-13}
  & &{Major Class} &{Root Class} &{Color} &{Position} &  &{SUC} & {P$_{\mathrm{Norm}}$} &{P} &{SUC} & {P$_{\mathrm{Norm}}$} &{P} \\
  \hline
  \ding{172}&{VLT}$_{\mathrm{SCAR}}^{\:t}$ &- &- &-  &- &$\surd$ &63.9  &73.3 &67.9  &49.8 &58.3 &51.1\\
  \ding{173}&{VLT}$_{\mathrm{SCAR}}^{\:t}$ &- &- &-  &- &- &53.4  &60.7 &54.6  &41.1 &49.1 &42.9\\
  \ding{174}&{VLT}$_{\mathrm{SCAR}}^{\:attr}$ &- &$\surd$ &- &-  &$\surd$      &65.5	&75.1 &71.1 &51.0 &57.5 &52.5 \\
  \ding{175}&{VLT}$_{\mathrm{SCAR}}^{\:attr}$ &- &$\surd$ &- &-  &-  &59.0	&66.5 &62.0 &47.0 &53.6 &47.8 \\


  \ding{176}&{VLT}$_{\mathrm{SCAR}}^{\:attr}$ &$\surd$ &$\surd$ &$\surd$  &- &$\surd$      &66.1	&74.8 &70.7 &{51.3} &57.8 &52.4 \\
  \ding{177}&{VLT}$_{\mathrm{SCAR}}^{\:attr}$ &$\surd$ &$\surd$ &$\surd$  &- &-  &57.8	&65.4 &60.9 &{46.4} &53.0 &47.5 \\


  \ding{178}&{VLT}$_{\mathrm{SCAR}}^{\:attr}$ &$\surd$ &$\surd$ &$\surd$ &$\surd$  &$\surd$      &66.4	&75.2 &72.0 &{51.8} &59.3 &53.1 \\
  \ding{179}&{VLT}$_{\mathrm{SCAR}}^{\:attr}$ &$\surd$ &$\surd$ &$\surd$ &$\surd$  &-  &57.1	&64.6 &60.5 &{45.7} &52.2 &46.3 \\


  \hline

  \ding{182}&{VLT}$_{\mathrm{TT}}^{\:t}$ &- &- &- &-  &$\surd$      &67.3  &78.0 &72.1  &53.1 &59.3 &53.3 \\
  \ding{183}&{VLT}$_{\mathrm{TT}}^{\:t}$ &- &- &- &-  &-   &61.0  &71.5 &63.4  &49.1 &55.5 &48.3 \\

  \ding{184}&{VLT}$_{\mathrm{TT}}^{\:attr}$ &$\surd$ &$\surd$ &$\surd$ &$\surd$  &$\surd$      &68.0	&79.6 &75.1 &{53.8} &59.9 &54.6 \\
  \ding{185}&{VLT}$_{\mathrm{TT}}^{\:attr}$ &$\surd$ &$\surd$ &$\surd$ &$\surd$  &-  &61.7	&73.4 &66.8 &{50.6} &56.8 &51.4 \\


  \hline
  \end{tabular}}
  \end{center}
  \end{table*}

\begin{table*}[!ht]
  \caption{Comparing different data volumes and sources for training with {sentence} annotation. ``SiamCAR Four Datasets'' consist of VID, Youtube-BB, DET and COCO, ``SiamCAR Seven Datasets'' consist of VID, Youtube-BB, DET, COCO, GOT-10K, LaSOT and TNL2K,  ``TransT Four Datasets'' consist of COCO, GOT-10K, LaSOT and TrackingNet.}
  \label{tab:dataabl}
  \begin{center}
  \newcommand{\dist}{\hspace{1pt}}
  \renewcommand\arraystretch{1.2}
  \resizebox{0.85\linewidth}{!}{
  \begin{tabular}{@{}c|c|c|c|ccc|ccc}
  \hline
  \multirow{2}{*}{\#} &\multirow{2}{*}{{Method}} &\multirow{2}{*}{{Data Volume}} &\multirow{2}{*}{{Data Source}} &\multicolumn{3}{c|}{{LaSOT}}  &\multicolumn{3}{c}{{TNL2K}}\\
  & & & &{SUC} & {P$_{\mathrm{Norm}}$}	&{P} &{SUC}	& {P$_{\mathrm{Norm}}$}	&{P}\\
  
  \hline
  \ding{172} &\tabincell{c}{SiamCAR} &\tabincell{c}{60W$\times$20Epoch} &\tabincell{c}{SiamCAR Four Datasets}
  &50.7  &60.0	&51.0	&35.3 &43.6	&38.4\\
  \ding{173} &\tabincell{c}{SiamCAR} &\tabincell{c}{60W$\times$20Epoch} &\tabincell{c}{LaSOT}
  &51.6 &61.6	&52.3	&35.0 &42.2	&36.4\\
  \ding{174} &\tabincell{c}{SiamCAR} &\tabincell{c}{120W$\times$20Epoch} &\tabincell{c}{SiamCAR Seven Datasets}
  &48.7 &57.6	&46.6	&39.7 &45.1	&39.2\\
  
  \hline
  \ding{175} &\tabincell{c}{{VLT$_{\mathbf{\mathrm{SCAR}}}$}} &\tabincell{c}{60W$\times$20Epoch} &\tabincell{c}{LaSOT}
  &57.0 &65.4	&58.6	&39.0 &47.6	&39.8\\
  \ding{176} &\tabincell{c}{{VLT$_{\mathbf{\mathrm{SCAR}}}$}} &\tabincell{c}{120W$\times$20Epoch} &\tabincell{c}{SiamCAR Seven Datasets}
  &63.9 &73.3	&67.9	&49.8 &58.3	&51.1\\
  
  \hline
  \ding{177} &\tabincell{c}{TransT} &\tabincell{c}{3.8W$\times$1000Epoch} &\tabincell{c}{TransT Four Datasets}
  &64.9 &73.8	&69.0	&50.7 &57.1	&51.7\\
  \ding{178} &\tabincell{c}{TransT} &\tabincell{c}{3.8W$\times$1000Epoch} &\tabincell{c}{TransT Four Datasets, TNL2K}
  &62.2 &70.1 &65.2 &51.2 &57.8 &52.3\\
  
  \hline
  \ding{179} &\tabincell{c}{{VLT$_{\mathbf{\mathrm{TT}}}$}} &\tabincell{c}{3.8W$\times$1000Epoch} &\tabincell{c}{TransT Four Datasets, TNL2K}
  &67.3 &78.0	&72.1	&53.1 &59.3	&53.3\\

  \hline
  \end{tabular}}
  \end{center}
  \end{table*}

\begin{table}[!t]
  \caption{Training with different volumes of language-annotated data on LaSOT.}
  \label{tab:analysis_supp:a}
  \begin{center}
  \newcommand{\dist}{\hspace{3pt}}
  \newcommand{\best}[1]{{\textcolor{red}{#1}}}
  \newcommand{\scnd}[1]{{\textcolor{blue}{#1}}}
  \renewcommand\arraystretch{1.2}
  
  \resizebox{0.75\linewidth}{!}{
  \begin{tabular}{c| c c c}
    \hline
         {Settings} & {SUC (\%)} &{P$_{\mathrm{Norm}}$ (\%)}  & {P (\%)} \\
    \hline
      ~~{50\%}~~  &58.9 &~66.3~  &61.4 \\
      ~~{75\%}~~ &~61.8~ &~72.1~  &~64.9~ \\
      ~~{100\%}~~ &~63.9~ &~73.3~  &~67.9~ \\
    \hline
\end{tabular}}
  \end{center}
  \end{table}

\textbf{Volume and Source of Training Data.} The quality of learned VL representation is deeply influenced by training data volume and source. As illustrated in Table~\ref{tab:dataabl}, we experiment by comparing VLT$^{\:t}_{\mathbf{\mathrm{SCAR}}}$, VLT$^{\:t}_{\mathbf{\mathrm{TT}}}$ and their baselines (\textit{i.e.}, SiamCAR and TransT) with different data settings:

(1) We retrain SiamCAR with the same data setting of VLT$_{\mathbf{\mathrm{SCAR}}}$ (\ding{174} \textit{v.s.} \ding{176}). Compared with the default setting (\ding{172}), double data volume and three more data sources contain different biases, which affect the trained model to produce biased outcomes, as illustrated in~\cite{rebuttal_1,rebuttal_2,rebuttal_3}. From the results (\ding{174}), the addition of TNL2K significantly improves the default \ding{172} with $4.4\%$ gains in SUC on TNL2K, whereas the performance on LaSOT slightly decreases. Compared to VLT$_{\mathbf{\mathrm{SCAR}}}$ with the same setting (\ding{176}), \ding{174} is still suppressed for $8.6\%/7.4\%$ of SUC and P on TNL2K, respectively.

(2) We also retrain VLT$_{\mathbf{\mathrm{SCAR}}}$ with the only LaSOT (\ding{175}), which keeps aligned with \ding{173}. The SUC scores on LaSOT and TNL2K degrade heavily to $57.0\%/39.0\%$ compared to the default \ding{176}, respectively. {This is caused by the great reduction of the language-annotated training data, i.e., from 1120 (LaSOT)+1300 (TNL2K)+9335 (GOT-10k) to 1120 (LaSOT).} Our model is hard to learn a good multimodal representation with the quite less language-annotated training data, which violates our intention. Even though, our VLT$_{\mathbf{\mathrm{SCAR}}}$ (\ding{175}) still outperforms the baseline SiamCAR (\ding{173}) for $5.4\%/4.0\%$ of SUC scores on LaSOT/TNL2K.

(3) TransT is also retrained with the same data setting as VLT$_{\mathbf{\mathrm{TT}}}$ (\ding{178} \textit{v.s.} \ding{179}). More data sources bring similar biases and influence the performance as SiamCAR, compared to default TransT (\ding{177}). Our VLT$_{\mathbf{\mathrm{TT}}}$ (\ding{179}) still achieves superior scores on both LaSOT and TNL2K.

\textbf{Volume of Language-Annotated Training Data.} As illustrated in Sec~\ref{implementation_details}, half training pairs of each epoch come from the datasets without language annotations, we further explore the influence of training with different volumes of language-annotated data based on VLT$^{\:t}_{\mathbf{\mathrm{SCAR}}}$ and the results are presented in Table~\ref{tab:analysis_supp:a}. The default setting is noted as ``100\%''. For settings of ``50\%'' and ``75\%'', the chopped part is replaced with the vision-only data to keep the training data volume. It shows that as the language-annotated training pairs reduced, the performance on LaSOT~\cite{LaSOT} gradually decreases ($63.9\% \rightarrow 61.8\% \rightarrow 58.9\%$ in SUC), demonstrating more language-annotated data helps improve model capacity of VL trackers.

\begin{table}[!t]
  \caption{Ablation on backbone branch and stage with contrastive loss.}
  \vspace{-5pt}
  \label{tab:clip_stage}
  \begin{center}
  \newcommand{\dist}{\hspace{3pt}}
  \newcommand{\best}[1]{{\textcolor{red}{#1}}}
  \newcommand{\scnd}[1]{{\textcolor{blue}{#1}}}
  \renewcommand\arraystretch{1.2}
  
  \resizebox{1\linewidth}{!}{
  \begin{tabular}{@{}c|c|cc|cccc|cc@{}c@{}}
  \hline
  \multirow{2}{*}{\#}& \multirow{2}{*}{{Method}} &\multicolumn{2}{c|}{{Backbone Branch}}   &\multicolumn{4}{c|}{{Backbone Stage}} &\multicolumn{2}{c@{}}{{LaSOT}}  \\
  \cline{3-10}
  & &{Template} &{Search} &{1st} &{2nd} &{3rd} & {4th} &{SUC} &{P} \\
  \hline
  \ding{172}&{VLT}$_{\mathrm{SCAR}}$  &- &- &- &- &- &- &65.5 &70.8 \\
  \ding{173}&{VLT}$_{\mathrm{SCAR}}$  &$\surd$ &- &$\surd$ &- &- &-   	&66.0  &71.1  \\
  \ding{174}&{VLT}$_{\mathrm{SCAR}}$  &$\surd$ &- &- &$\surd$ &- &-   	&64.5  &69.4  \\
  \ding{175}&{VLT}$_{\mathrm{SCAR}}$  &$\surd$ &- &- &- &$\surd$ &-   	&63.7  &68.1  \\
  \ding{176}&{VLT}$_{\mathrm{SCAR}}$  &$\surd$ &- &- &- &- &$\surd$  	&63.0  &67.1  \\
  \ding{177}&{VLT}$_{\mathrm{SCAR}}$  &- &$\surd$ &$\surd$ &- &- &-  	&66.1  &71.0  \\ 
  \ding{178}&{VLT}$_{\mathrm{SCAR}}$  &$\surd$ &$\surd$ &$\surd$ &- &- &-  	&66.6  &71.4  \\

  \hline
  \end{tabular}}
  \end{center}
  \end{table}

\begin{table*}[t!]
	\centering
	\caption{Evaluating different settings on LaSOT: (a) the influence of symmetrical and our asymmetrical design, (b) adopting fixed ShuffleNet block or searching the post-processing block in ModaMixer, and (c) removing the residual connection (dubbed as ``Res'') of ModaMixer.}
	\label{tab:analysis}
	\newcommand{\best}[1]{{\textcolor{red}{#1}}}
	\newcommand{\scnd}[1]{{\textcolor{blue}{#1}}}
	\renewcommand\arraystretch{1.2}
\begin{minipage}[c]{.33\textwidth}
\center
{(a)}\\ \vspace{3pt}
\label{tab:analysis:a}
\setlength{\tabcolsep}{5pt}
\small
\begin{tabular}{c| c c}
    \hline
         {Settings} & {SUC}  & {P} \\
    \hline
      ~~{symmetrical}~~  &60.0  &61.7 \\
      ~~{asymmetrical}~~ &~63.9~  &~67.9~ \\
    \hline
\end{tabular}
\end{minipage}\hfill
\begin{minipage}[c]{.33\textwidth}
\center
{(b)}\\ \vspace{3pt}
\label{tab:analysis:b}
\setlength{\tabcolsep}{2pt}
\small
\begin{tabular}{c| c c}
    \hline
         {Settings} & {SUC}  & {P} \\
    \hline
      ~{Shuffle-ModaMixer}~ &~59.1~  &~62.2~ \\
      ~{NAS-ModaMixer}~  &~63.9~  &~67.9~ \\
    \hline
\end{tabular}
\end{minipage}\hfill
\begin{minipage}[c]{.33\textwidth}
\center
{(c)}\\ \vspace{3pt}
\label{tab:analysis:c}
\setlength{\tabcolsep}{2pt}
\small
\begin{tabular}{c| c c}
    \hline
         {Settings} & {SUC}  & {P} \\
    \hline
      {w/o. Res}  &~61.1~  &~63.6~ \\
      {w/. Res}  &~63.9~  &~67.9~ \\
    \hline
\end{tabular}
\end{minipage}
\end{table*}

\begin{table}[!t]
  \caption{Influence of different language models on LaSOT.}
  \vspace{-5pt}
  \label{tab:analysis_supp:b}
  \begin{center}
  \newcommand{\dist}{\hspace{3pt}}
  \newcommand{\best}[1]{{\textcolor{red}{#1}}}
  \newcommand{\scnd}[1]{{\textcolor{blue}{#1}}}
  \renewcommand\arraystretch{1.2}
  
  \resizebox{0.8\linewidth}{!}{
  \begin{tabular}{c| c c c}
    \hline
         {Settings} & {SUC (\%)} &{P$_{\mathrm{Norm}}$ (\%)}  & {P (\%)} \\
    \hline
      ~~{GPT-2}~\cite{gpt}~~ &~59.3~ &~67.1~  &~62.3~ \\
      ~~{BERT}~\cite{bert}~~  &~63.9~ &~73.3~  &~67.9~ \\
    \hline
\end{tabular}}
  \end{center}
  \end{table}

\subsubsection{Model Design}
\label{futher_ana:b}

\textbf{Contrastive Loss in Different Tracking Branches and Backbone Stages.} As shown in Fig.~\ref{fig:modamixer} and Fig.~\ref{fig:framework}, contrastive loss could be applied into the ModaMixer of different tracking branches and backbone stages. As shown in Table~\ref{tab:clip_stage}, we first ablate the contrastive loss at different representation learning stages by only operating on the template branch. The results show that applying contrastive loss at the first backbone stage achieves the best $66.0\%$ SUC on LaSOT (\ding{173} \textit{v.s.} \ding{174},\ding{175},\ding{176}). Conversely, applying contrastive loss in later stages even degrade the performance (\ding{174},\ding{175},\ding{176} \textit{v.s.} \ding{172}), which demonstrates later multimodal alignment would break the consistency of representation learning and result in the inferior performance.  

We then analyze the influence of contrastive loss applied to different tracking branches. Results in Table~\ref{tab:clip_stage} show that we can obtain consistent improvements whether integrating the loss to template or search branches (\ding{173},\ding{177} \textit{v.s.} \ding{172}), which evidences the effectiveness and generality of learning contrastive VL representation. When applying the loss to both template and search branches, the tracking performances are further enhanced with $66.6\%$ SUC of LaSOT (\ding{178}).

\textbf{Symmetric or Asymmetric?} The proposed asymmetric searching strategy is essential for achieving an adaptive vision-language representation. As illustrated in Table~\ref{tab:analysis:a}, we experiment by searching for a symmetric network (including both backbone and $\operatorname{Block_{ASearch}}$ in the ModaMixer) based on VLT$^{\:t}_{\mathbf{\mathrm{SCAR}}}$, but it is inferior to the asymmetric counterpart for $3.9\%/6.2\%$ of success rate (SUC) and precision (P) on LaSOT~\cite{LaSOT}, respectively, which empirically proves our argument.

\textbf{Asymmetry in ModaMixer.} The asymmetry is used in not only the backbone network, but also the ModaMixer. In our work, the post-processing layers for different signals (visual and mixed features) are decided by ASearch, which enables the adaption at both semantic levels (\ie, network depth) and different input signals (\ie, template and search, pure-vision and mixed feature in each ModaMixer). As in Table~\ref{tab:analysis:b}, when replacing the post-processing layers with a fixed ShuffleNet block from SPOS~\cite{spos} (\ie, inheriting structure and weights from the last block in each backbone stage), the performance of VLT$^{\:t}_{\mathbf{\mathrm{SCAR}}}$ drops from $63.9\%$ to 59.1\% in SUC on LaSOT. This reveals that the proposed ASearch is important for building a better VL learner.

\textbf{Residual Connection of ModaMixer} The residual connection~\cite{resnet} is a commonly used trick to avoid information loss. In our VL representation learning, it provides more vision messages for better multimodal fusion. We experiment by removing the structure based on VLT$_{\mathbf{\mathrm{SCAR}}}^{\:t}$ as shown in Table~\ref{tab:analysis:c}. Compared to the default setting (\textit{i.e.,} w/. Res), the loss of additional vision details brings decreases for $2.7\%/2.1\%$ of SUC on LaSOT/TNL2K, respectively. Even though, the performance is still much higher than the baseline. This demonstrates the improvements are mainly attributed to the multimodal fusion.

\begin{table}[!t]
   \caption{Results of applying ModaMixer and ASearch to SiamRPN++.}
   \label{tab:siamrpn}
   \begin{center}
   \newcommand{\dist}{\hspace{1pt}}
   \renewcommand\arraystretch{1.2}
   \resizebox{1\linewidth}{!}{
   \begin{tabular}{c|ccc|ccc}
   \hline
   \multirow{2}{*}{{Method}} &\multicolumn{3}{c|}{{LaSOT}}  &\multicolumn{3}{c}{{TNL2K}}\\
  \cline{2-7}
   &{SUC} &{P$_{\mathrm{Norm}}$}	&{P} &{SUC}	&{P$_{\mathrm{Norm}}$}	&{P}\\
   
  \hline
  \tabincell{c}{{VLT$_{\mathbf{\mathrm{RPN++}}}$}}
  &59.0  &68.4 &62.6
   &45.8  &54.2 &47.4\\
  
  \tabincell{c}{{SiamRPN++}}
  &49.6  &56.9 &49.1
   &41.3  &48.2 &41.2\\
   
   \cline{1-7}
   \hline
   \end{tabular}}
   \end{center}
   \end{table}

\begin{table*}[!t]
	\centering
	\caption{Pipeline comparison of ASearch and LightTrack in terms of time complexity.}
	\label{tab:comporison}
	\newcommand{\best}[1]{{\textcolor{red}{#1}}}
	\newcommand{\scnd}[1]{{\textcolor{blue}{#1}}}
	\renewcommand\arraystretch{1.2}
\resizebox{0.75\linewidth}{!}{
\begin{tabular}{c| c| c}
    \hline
         {Steps} & {LightTrack}~\cite{yan2021lighttrack}  & {ASearch in VLT} (Ours) \\
    \hline
      {1st step} &\tabincell{c}{{Pretraining} backbone supernet\\on ImageNet~\cite{imagenet}} &\tabincell{c}{{Reusing} trained backbone\\supernet of SPOS~\cite{spos}}\\
      \hline
      {2nd step} &\tabincell{c}{Training tracking supernet\\on tracking datasets} &\tabincell{c}{Training tracking supernet\\on tracking datasets}\\
      \hline
      {3rd step} &\tabincell{c}{Searching with evolutionary\\algorithm on tracking supernet} &\tabincell{c}{Searching with evolutionary\\algorithm on tracking supernet}\\
      \hline
      {4th step} &\tabincell{c}{{Retraining} searched backbone\\subset on ImageNet~\cite{imagenet}} &\tabincell{c}{{Reusing} trained backbone\\supernet of SPOS~\cite{spos}}\\
      \hline
      {5th step} &\tabincell{c}{Finetuning searched tracking\\subset on tracking datasets} &\tabincell{c}{Finetuning searched tracking\\subset on tracking datasets}\\
      \hline
      {Network searching cost} & {$\sim$40 Tesla-V100} GPU days & {$\sim$3 RTX-2080Ti} GPU days\\
    \hline
\end{tabular}}
\end{table*}

\textbf{Different Language Models.} As described in Sec.~\ref{sec:modalixer}, the language model of BERT~\cite{bert} is adopted to abstract the semantics of the sentence, which directly relates to the learning of vision-language representation. To show the influence of different language models, we compare the results of using BERT~\cite{bert} and GPT-2~\cite{gpt} based on VLT$^{\:t}_{\mathbf{\mathrm{SCAR}}}$, as shown in Table~\ref{tab:analysis_supp:b}. An interesting finding is that GPT-2~\cite{gpt} even decreases the performances, which is discrepant with recent studies in natural language processing. One possible reason is that the bi-directional learning strategy in BERT~\cite{bert} can better capture the context information of a sentence than the self-regression in GPT-2~\cite{gpt}.

\textbf{Multimodal Vision-Language Tracking with SiamRPN++~\cite{Siamrpn++}.} We apply our method to another pure-CNN-based tracker SiamRPN++ (the baseline of current best VL tracker SNLT~\cite{SNLT}) and the results are shown in Table~\ref{tab:siamrpn}. Compared with the baseline SiamRPN++, the new tracker (dubbed VLT$_{\mathbf{\mathrm{RPN++}}}$) achieves considerable SUC gains of $9.4\%/4.5\%$ on LaSOT/TNL2K, respectively. This demonstrates the effectiveness of multimodal representation learning (ModaMixer) and the proposed ASearch.

\textbf{Comparison of ASearch and LightTrack~\cite{yan2021lighttrack}.} Despite greatly boosting the tracking performance, Neural Architecture Search (NAS) brings complicated training processes and large computation costs. Considering the complexity, we ease unnecessary steps of ASearch to achieve a better trade-off between training time and performance. Taking another NAS-based tracker (\ie, LightTrack~\cite{yan2021lighttrack}) as the comparison, we demonstrate the efficiency of our proposed ASearch. As illustrated in Table~\ref{tab:comporison}, NAS-based trackers usually need to first pretrain the supernet on ImageNet~\cite{imagenet} to initialize the parameters, which results in high time complexity in training. LightTrack even trains the backbone network on ImageNet~\cite{imagenet} twice (\ie, the 1st and 4th steps), which heavily increases the time complexity. By contrast, our ASearch avoids this cost by reusing the pre-trained supernet from SPOS, which is much more efficient.

\begin{figure*}[!t]
\centering
\begin{minipage}[t]{0.8\textwidth}
\subfloat[]{\includegraphics[width=0.76\textwidth]{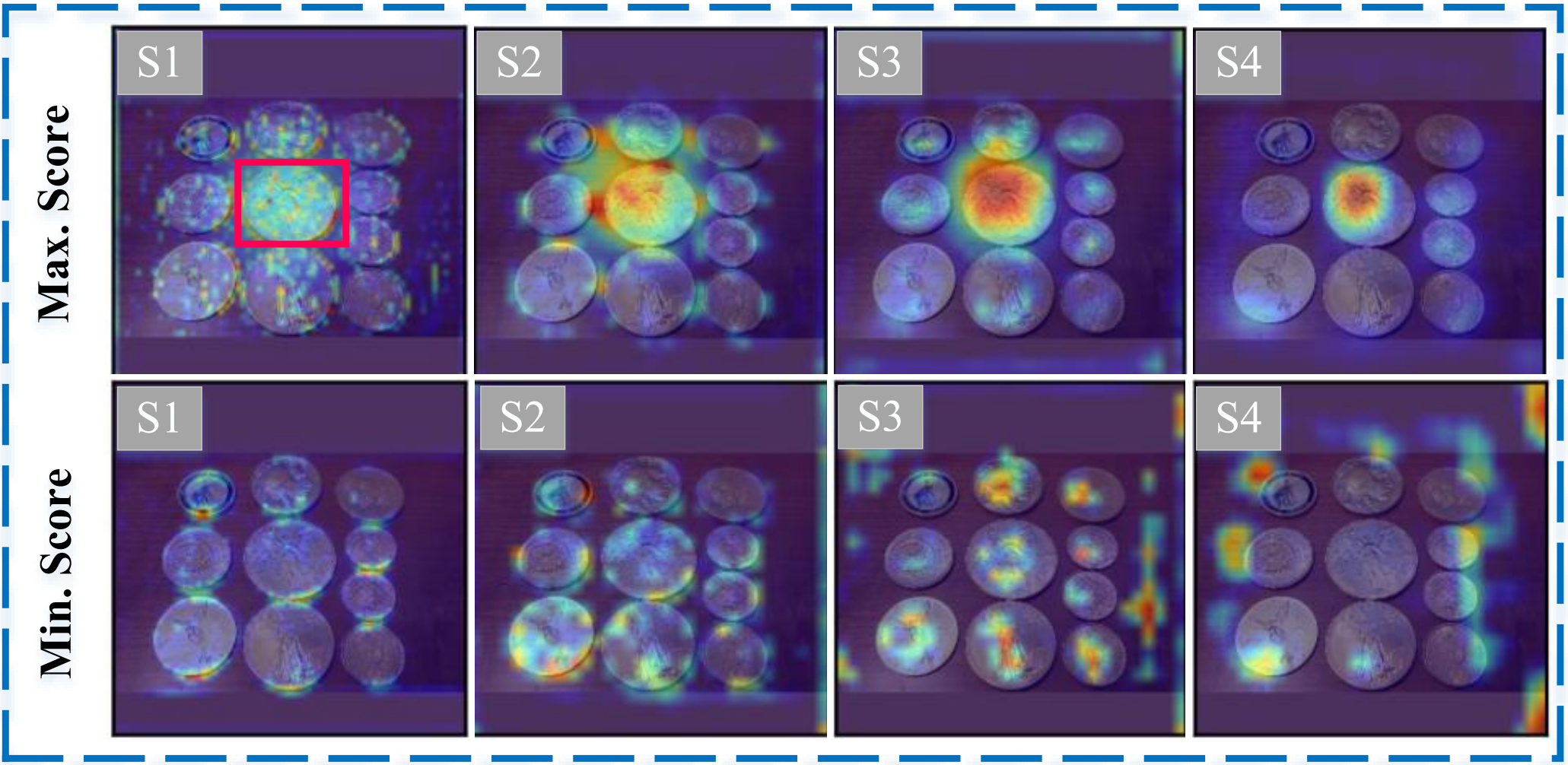}
\label{fig:vis_1}}
\hfill
\subfloat[]{\includegraphics[width=0.235\textwidth]{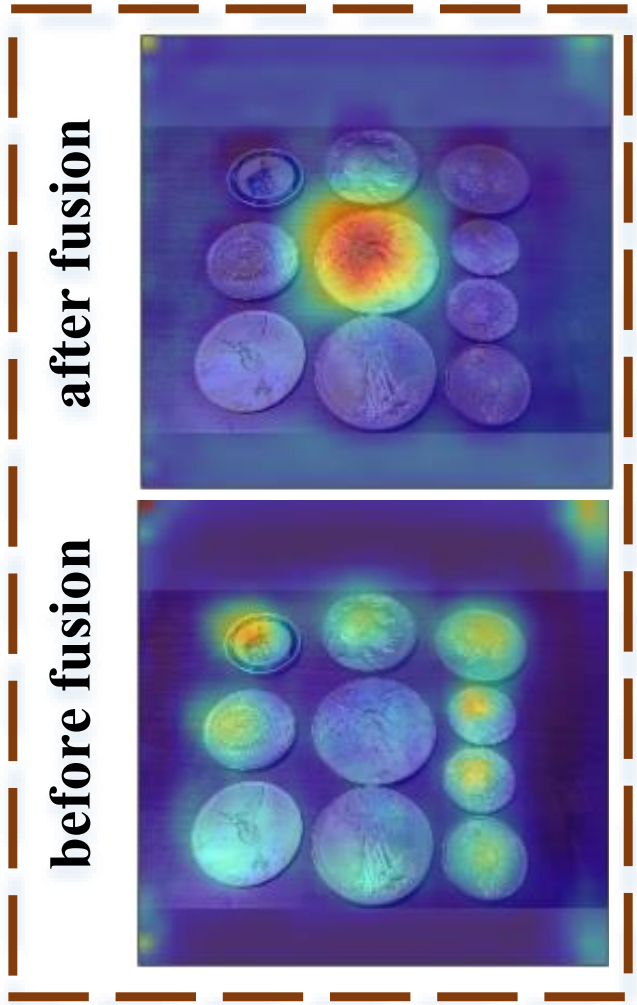}
\label{fig:vis_2}}
\end{minipage}
\caption{(a) feature channel with maximum/minimum (top/bottom) selection scores from ModaMixer in stage1-4. (b) activation map before/after (top/bottom) multimodal fusion in ModaMixer.}
\label{fig:vis}
\end{figure*}

\begin{figure*}[!t]
\centering
\begin{minipage}[t]{0.36\textwidth}
    \centering
	\subfloat[][Comparison with different trackers.]{\includegraphics[width = \textwidth]{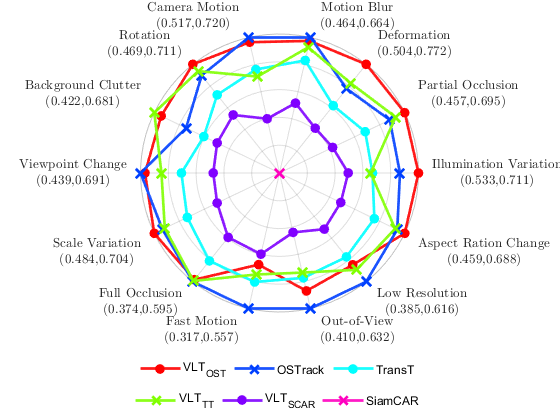}\label{fig:attribute}}
\end{minipage}
\;\;\;\;\;\;\;\;\;\;\;\;\;\;
\begin{minipage}[t]{0.36\textwidth}
    \centering
 \subfloat[][Ablation on components of VLT$_{\mathrm{SCAR}}$.]{\includegraphics[width = \textwidth]{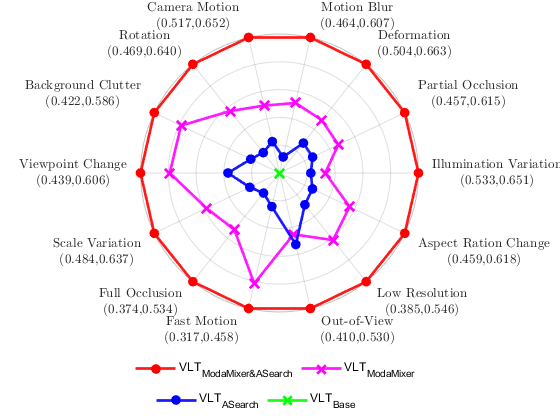}\label{fig:attribute_ablation}}
\end{minipage}
\caption{AUC scores of different attributes on LaSOT.}
\label{fig:lasotatt}
\vspace{-5pt}
\end{figure*}

\subsubsection{Visualization}
\label{futher_ana:c}

\textbf{Channel Selection by ModaMixer.} ModaMixer translates the language description to a channel selector to reweight visual features. As shown in Fig.~\ref{fig:vis}, the channel activation maps with maximum selection scores usually correspond to the target, while the surrounding distractors are successfully assigned with minimum scores (Fig.~\ref{fig:vis} (a)-bottom). Besides, with multimodal fusion (or channel selection), the network can enhance the response of target and meanwhile suppress the distractors (see Fig.~\ref{fig:vis} (b)). This evidences our argument that language embedding can help identify semantics in visual feature channels and effectively select useful information for localizing targets.

\begin{figure*}[!t]
\center
  \begin{minipage}{0.895\linewidth}
  \centerline{\includegraphics[width=\textwidth]{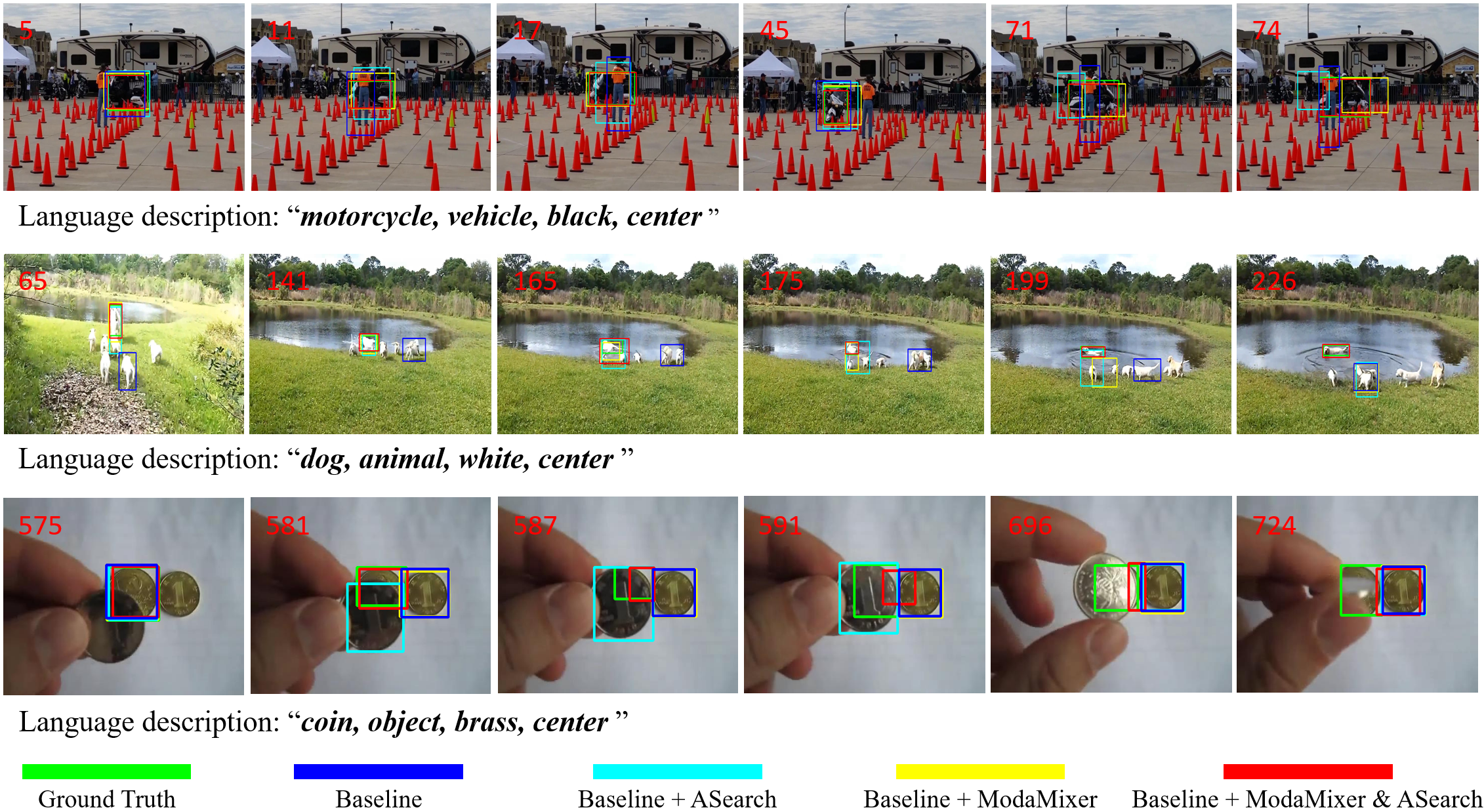}}
  \end{minipage}
   
  \caption{The first two rows show the success of our tracker in locating target object in complex scenarios, while the third row exhibits a failure case of our method when the target is occluded for a long period (with around 100 frames).}
  \label{fig:vis_supp}
  \end{figure*}

\begin{figure*}[!t]
\center
  \begin{minipage}{0.895\linewidth}
  \centerline{\includegraphics[width=\textwidth]{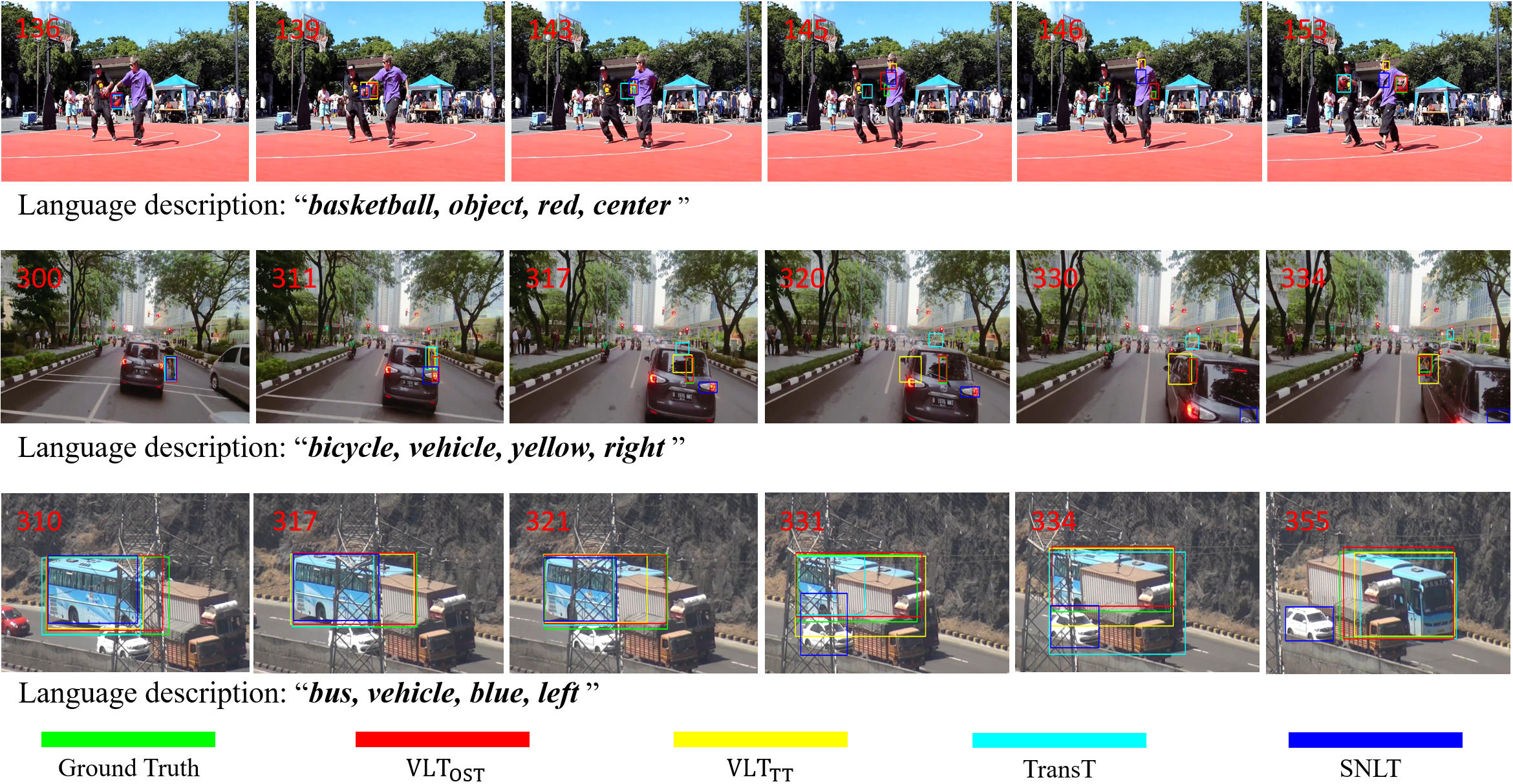}}
  \end{minipage}
   
  
  \caption{Results visualization of different trackers. The comparison shows that our VLT$_{\mathrm{OST}}$ could perform robust tracking under complex scenarios (\textit{e.g.,} Deformation, Disappearance, Occlusion and Similar Interferences).}
  \label{fig:qualitative}
  \vspace{-4pt}
  \end{figure*}

\textbf{Detailed Performance Analysis.} Fig.~\ref{fig:lasotatt} presents the attribute-based evaluation on LaSOT~\cite{LaSOT}. We compare the proposed VLT$_{\mathrm{SCAR}}$, VLT$_{\mathrm{TT}}$ and VLT$_{\mathrm{OST}}$ with representative state-of-the-art algorithms, as shown in Fig.~\ref{fig:attribute}. It shows that our methods are more effective than other competing trackers on most attributes, proving the effectiveness of high-level linguistic messages. Fig.~\ref{fig:attribute_ablation} shows the ablation on different components of VLT$_{\mathrm{SCAR}}$, which evidences that the integration of ModaMixer and ASearch is necessary for a powerful multimodal vision-language tracker.

\textbf{Visualization of Tracking Result and Failure Case.} As shown in Fig.~\ref{fig:vis_supp}, the proposed VLT$_{\mathrm{SCAR}}$ delivers more robust tracking under deformation, occlusion (the first row) and interference with similar objects (the second row). It demonstrates the effectiveness of learned multimodal representation, especially in complex environments. The third row shows the failure case of our tracker. In this case, the target is fully occluded for about 100 frames and distracted by similar objects, leading to ineffectiveness of our tracker in learning helpful information. A possible solution to deal with this is to apply a global searching strategy, and we leave this to future work. Fig.~\ref{fig:qualitative} shows that our VLT$_{\mathrm{OST}}$ achieves the best performance compared to other SOTAs. It demonstrates the resilience of our tracker and effectiveness of proposed multimodal vision-language tracking in complex environments.

\begin{figure*}[!t]
\center
  \begin{minipage}{0.895\linewidth}
  \centerline{\includegraphics[width=\textwidth]{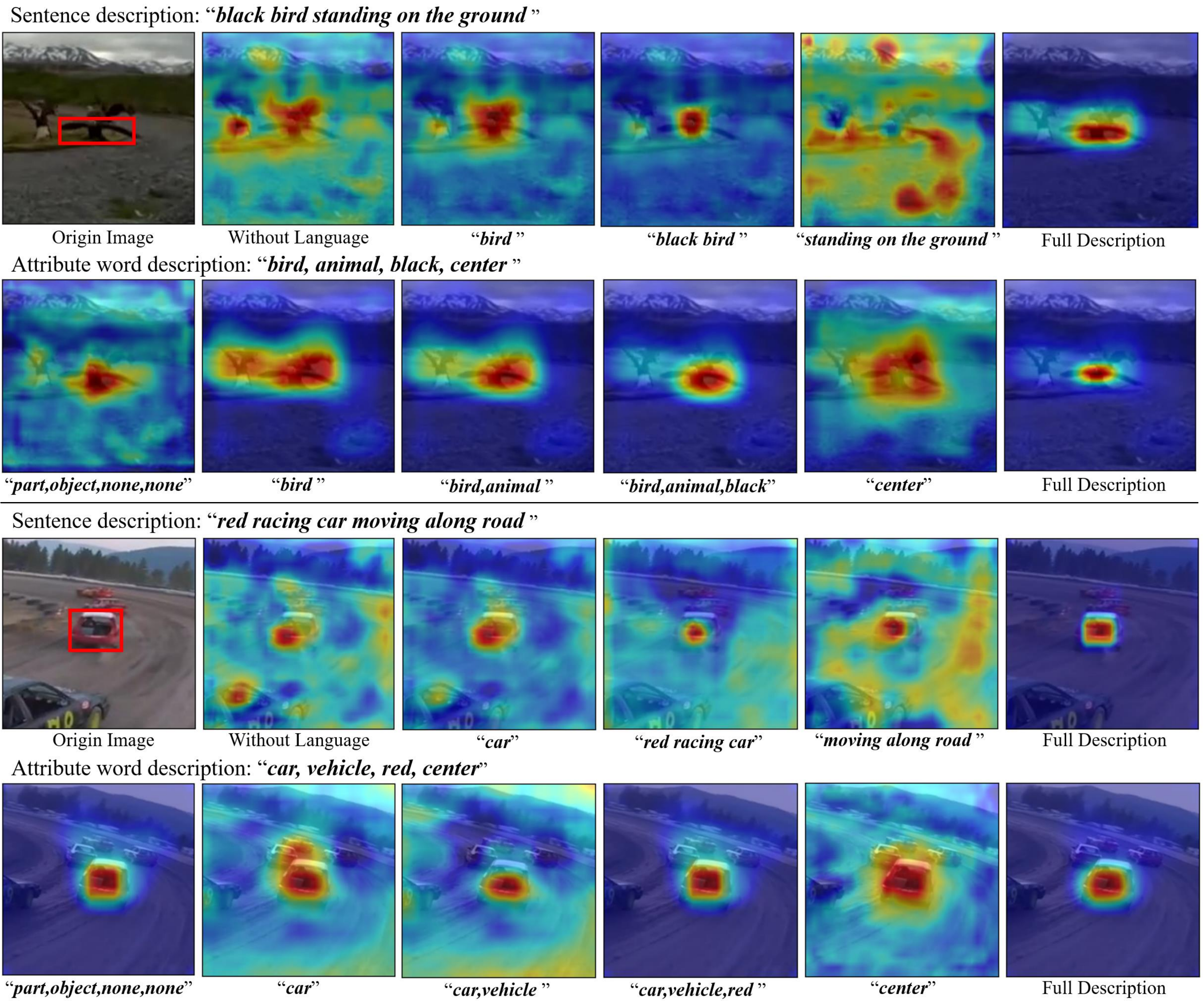}}
  \end{minipage}
   
  \caption{Activation visualization of VLT$_{\mathrm{SCAR}}$ with different language descriptions (\ie, sentence and attribute words) using GradCAM~\cite{gradcam}. The high-level language description endows our VL tracker the distinguishability between the target and interferences.}
  \label{fig:cam}
  \vspace{-4pt}
  \end{figure*}

\textbf{Activation Analysis of Different Language Descriptions} Language description provides high-level semantics to enhance the target-specific channels while suppressing the target-irrelevant ones. As presented in Fig.~\ref{fig:cam}, we show the effects of different words in both sentence and attribute annotations to evidence the enhanced discernibility by language. In the first row using sentence annotation, the second column shows that the VLT$_{\mathrm{SCAR}}$ without language description focuses on two birds (red areas), interfered by the same object class. When introducing the word ``bird'', the response of the similar object is obviously suppressed. With a more detailed ``black bird'', the responses of distractors almost disappear, which reveals that more specific annotation can help the tracker better locate the target area. Furthermore, we also try to track the target with only environmental description, \ie, ``standing on the ground''. The result in column 5 shows that the background is enhanced while the target area is suppressed. The comparison evidences that the language description of object class is crucial for the tracker to distinguish the target from the background clutter, while the mere description of the environment may introduce interference instead. The last column shows the activation maps with full description, where the tracker can precisely locate the target, demonstrating the effectiveness of the learned unified-adaptive vision-language representation. We observe similar improvements by more general attribute words on distinguishing the target from irrelevant interferences (\eg, the second row).

\section{Conclusion}
In this work, we unleash the immense potential of multimodal Vision-language (VL) tracking by addressing two main limitations: insufficient VL data with limited sentence annotation and ineffective VL interaction learning. First, we establish a large-scale VL database with comprehensive attribute annotations, enabling more robust and diverse training. Second, we propose a unified-adaptive VL representation framework, consisting of our ModaMixer and asymmetric networks, to effectively learn and integrate multimodal information. In experiments, our approach generally and effectively improves trackers under different tracking frameworks, and reaches new SOTA levels on six challenging tracking benchmarks. Besides, we provide a theoretical explanation to evidence the effectiveness of our method. We believe that our work opens up new possibilities for future tracking research, particularly in leveraging diversified multimodal messages.


%

\ifCLASSOPTIONcaptionsoff
  \newpage
\fi



%


\bibliographystyle{IEEEtran}
\bibliography{references.bib}

\begin{thebibliography}{10}
\providecommand{\url}[1]{#1}
\csname url@samestyle\endcsname
\providecommand{\newblock}{\relax}
\providecommand{\bibinfo}[2]{#2}
\providecommand{\BIBentrySTDinterwordspacing}{\spaceskip=0pt\relax}
\providecommand{\BIBentryALTinterwordstretchfactor}{4}
\providecommand{\BIBentryALTinterwordspacing}{\spaceskip=\fontdimen2\font plus
\BIBentryALTinterwordstretchfactor\fontdimen3\font minus
  \fontdimen4\font\relax}
\providecommand{\BIBforeignlanguage}[2]{{%
\expandafter\ifx\csname l@#1\endcsname\relax
\typeout{** WARNING: IEEEtran.bst: No hyphenation pattern has been}%
\typeout{** loaded for the language `#1'. Using the pattern for}%
\typeout{** the default language instead.}%
\else
\language=\csname l@#1\endcsname
\fi
#2}}
\providecommand{\BIBdecl}{\relax}
\BIBdecl

\bibitem{SiamCAR}
D.~Guo, J.~Wang, Y.~Cui, Z.~Wang, and S.~Chen, ``{SiamCAR:} {S}iamese fully
  convolutional classification and regression for visual tracking,'' in
  \emph{CVPR}, 2020.

\bibitem{ostrack}
B.~Ye, H.~Chang, B.~Ma, S.~Shan, and X.~Chen, ``Joint feature learning and
  relation modeling for tracking: A one-stream framework,'' in \emph{ECCV},
  2022.

\bibitem{TransT}
X.~Chen, B.~Yan, J.~Zhu, D.~Wang, X.~Yang, and H.~Lu, ``Transformer tracking,''
  in \emph{CVPR}, 2021.

\bibitem{vlt}
M.~Guo, Z.~Zhang, H.~Fan, and L.~Jing, ``Divert more attention to
  vision-language tracking,'' \emph{NeurIPS}, 2022.

\bibitem{vlvqa1}
K.~Uehara, N.~Duan, and T.~Harada, ``Learning to ask informative sub-questions
  for visual question answering,'' in \emph{CVPR}, 2022.

\bibitem{vlvqa2}
M.~Zhao, B.~Li, J.~Wang, W.~Li, W.~Zhou, L.~Zhang, S.~Xuyang, Z.~Yu, X.~Yu,
  G.~Li \emph{et~al.}, ``Towards video text visual question answering:
  benchmark and baseline,'' \emph{NeurIPS}, 2022.

\bibitem{vlcaption1}
J.~Gamper and N.~Rajpoot, ``Multiple instance captioning: Learning
  representations from histopathology textbooks and articles,'' in \emph{CVPR},
  2021.

\bibitem{vlcaption2}
W.~Zhao, X.~Wu, and J.~Luo, ``Multi-modal dependency tree for video
  captioning,'' \emph{NeurIPS}, 2021.

\bibitem{detsurvey}
S.~S.~A. Zaidi, M.~S. Ansari, A.~Aslam, N.~Kanwal, M.~Asghar, and B.~Lee, ``A
  survey of modern deep learning based object detection models,'' \emph{Digital
  Signal Processing}, 2022.

\bibitem{segsurvey}
S.~Minaee, Y.~Boykov, F.~Porikli, A.~Plaza, N.~Kehtarnavaz, and D.~Terzopoulos,
  ``Image segmentation using deep learning: A survey,'' \emph{T-PAMI}, 2021.

\bibitem{clip}
A.~Radford, J.~W. Kim, C.~Hallacy, A.~Ramesh, G.~Goh, S.~Agarwal, G.~Sastry,
  A.~Askell, P.~Mishkin, J.~Clark \emph{et~al.}, ``Learning transferable visual
  models from natural language supervision,'' in \emph{ICML}, 2021.

\bibitem{sam}
A.~Kirillov, E.~Mintun, N.~Ravi, H.~Mao, C.~Rolland, L.~Gustafson, T.~Xiao,
  S.~Whitehead, A.~C. Berg, W.-Y. Lo \emph{et~al.}, ``Segment anything,''
  \emph{arXiv}, 2023.

\bibitem{visionllm}
W.~Wang, Z.~Chen, X.~Chen, J.~Wu, X.~Zhu, G.~Zeng, P.~Luo, T.~Lu, J.~Zhou,
  Y.~Qiao \emph{et~al.}, ``Visionllm: Large language model is also an
  open-ended decoder for vision-centric tasks,'' \emph{arXiv}, 2023.

\bibitem{SNLT}
Q.~Feng, V.~Ablavsky, Q.~Bai, and S.~Sclaroff, ``Siamese natural language
  tracker: Tracking by natural language descriptions with siamese trackers,''
  in \emph{CVPR}, 2021.

\bibitem{Li}
Z.~Li, R.~Tao, E.~Gavves, C.~G. Snoek, and A.~W. Smeulders, ``Tracking by
  natural language specification,'' in \emph{CVPR}, 2017.

\bibitem{Feng}
Q.~Feng, V.~Ablavsky, Q.~Bai, G.~Li, and S.~Sclaroff, ``Real-time visual object
  tracking with natural language description,'' in \emph{WACV}, 2020.

\bibitem{OTB2015}
Y.~Wu, J.~Lim, and M.~H. Yang, ``Object tracking benchmark,'' \emph{IEEE
  T-PAMI}, 2015.

\bibitem{trackingnet}
M.~Muller, A.~Bibi, S.~Giancola, S.~Alsubaihi, and B.~Ghanem, ``Tracking{N}et:
  A large-scale dataset and benchmark for object tracking in the wild,'' in
  \emph{ECCV}, 2018.

\bibitem{LaSOT}
H.~Fan, L.~Lin, F.~Yang, P.~Chu, G.~Deng, S.~Yu, H.~Bai, Y.~Xu, C.~Liao, and
  H.~Ling, ``{LaSOT}: A high-quality benchmark for large-scale single object
  tracking,'' in \emph{CVPR}, 2019.

\bibitem{GOT10K}
L.~Huang, X.~Zhao, and K.~Huang, ``Got-10k: A large high-diversity benchmark
  for generic object tracking in the wild,'' \emph{IEEE T-PAMI}, 2019.

\bibitem{TNL2K}
X.~Wang, X.~Shu, Z.~Zhang, B.~Jiang, Y.~Wang, Y.~Tian, and F.~Wu, ``Towards
  more flexible and accurate object tracking with natural language: Algorithms
  and benchmark,'' in \emph{CVPR}, 2021.

\bibitem{pnas_causal}
S.~Badde, F.~Hong, and M.~S. Landy, ``Causal inference and the evolution of
  opposite neurons,'' \emph{PNAS}, 2021.

\bibitem{LaSOT_Extention}
H.~Fan, H.~Bai, L.~Lin, F.~Yang, P.~Chu, G.~Deng, S.~Yu, M.~Huang, J.~Liu,
  Y.~Xu \emph{et~al.}, ``Lasot: A high-quality large-scale single object
  tracking benchmark,'' \emph{IJCV}, 2021.

\bibitem{channelmeaning1}
X.~Gao, Y.~Zhao, {\L}.~Dudziak, R.~Mullins, and C.-z. Xu, ``Dynamic channel
  pruning: Feature boosting and suppression,'' \emph{arXiv}, 2018.

\bibitem{channelmeaning2}
B.~Yang, J.~Yan, Z.~Lei, and S.~Z. Li, ``Convolutional channel features,'' in
  \emph{ICCV}, 2015.

\bibitem{nas1}
B.~Zoph and Q.~V. Le, ``Neural architecture search with reinforcement
  learning,'' in \emph{OCLR}, 2017.

\bibitem{nas2}
E.~Real, A.~Aggarwal, Y.~Huang, and Q.~V. Le, ``Regularized evolution for image
  classifier architecture search,'' in \emph{AAAI}, 2019.

\bibitem{imagenet}
J.~Deng, W.~Dong, R.~Socher, L.-J. Li, K.~Li, and L.~Fei-Fei, ``Imagenet: A
  large-scale hierarchical image database,'' in \emph{CVPR}, 2009.

\bibitem{MOSSE}
D.~S. Bolme, J.~R. Beveridge, B.~A. Draper, and Y.~M. Lui, ``Visual object
  tracking using adaptive correlation filters,'' in \emph{CVPR}.\hskip 1em plus
  0.5em minus 0.4em\relax IEEE, 2010.

\bibitem{KCF}
J.~F. Henriques, R.~Caseiro, P.~Martins, and J.~Batista, ``High-speed tracking
  with kernelized correlation filters,'' \emph{IEEE T-PAMI}, 2014.

\bibitem{ECO}
M.~Danelljan, G.~Bhat, F.~Shahbaz~Khan, and M.~Felsberg, ``Eco: Efficient
  convolution operators for tracking,'' in \emph{CVPR}, 2017.

\bibitem{Atom}
M.~Danelljan, G.~Bhat, F.~S. Khan, and M.~Felsberg, ``{ATOM: Accurate} tracking
  by overlap maximization,'' in \emph{CVPR}, 2019.

\bibitem{prdimp}
M.~Danelljan, L.~V. Gool, and R.~Timofte, ``Probabilistic regression for visual
  tracking,'' in \emph{CVPR}, 2020.

\bibitem{Siamfc}
L.~Bertinetto, J.~Valmadre, J.~F. Henriques, A.~Vedaldi, and P.~H.~S. Torr,
  ``Fully-convolutional siamese networks for object tracking,'' in
  \emph{ECCVW}, 2016.

\bibitem{tao2016siamese}
R.~Tao, E.~Gavves, and A.~W. Smeulders, ``Siamese instance search for
  tracking,'' in \emph{CVPR}, 2016.

\bibitem{Siamrpn}
B.~Li, J.~Yan, W.~Wu, Z.~Zhu, and X.~Hu, ``High performance visual tracking
  with siamese region proposal network,'' in \emph{CVPR}, 2018.

\bibitem{SiamDW}
Z.~Zhang and H.~Peng, ``Deeper and wider siamese networks for real-time visual
  tracking,'' in \emph{CVPR}, 2019.

\bibitem{Siamrpn++}
B.~Li, W.~Wu, Q.~Wang, F.~Zhang, J.~Xing, and J.~Yan, ``Siamrpn++: Evolution of
  siamese visual tracking with very deep networks,'' in \emph{CVPR}, 2019.

\bibitem{C-RPN}
H.~Fan and H.~Ling, ``Siamese cascaded region proposal networks for real-time
  visual tracking,'' in \emph{CVPR}, 2019.

\bibitem{Ocean}
Z.~Zhang, H.~Peng, J.~Fu, B.~Li, and W.~Hu, ``Ocean: Object-aware anchor-free
  tracking,'' in \emph{ECCV}, 2020.

\bibitem{Transformer}
A.~Vaswani, N.~Shazeer, N.~Parmar, J.~Uszkoreit, L.~Jones, A.~N. Gomez,
  {\L}.~Kaiser, and I.~Polosukhin, ``Attention is all you need,'' in
  \emph{NIPS}, 2017.

\bibitem{TrDiMP}
N.~Wang, W.~Zhou, J.~Wang, and H.~Li, ``Transformer meets tracker: Exploiting
  temporal context for robust visual tracking,'' in \emph{CVPR}, 2021.

\bibitem{stark}
B.~Yan, H.~Peng, J.~Fu, D.~Wang, and H.~Lu, ``Learning spatio-temporal
  transformer for visual tracking,'' in \emph{ICCV}, 2021.

\bibitem{mixformer}
Y.~Cui, J.~Cheng, L.~Wang, and G.~Wu, ``Mixformer: End-to-end tracking with
  iterative mixed attention,'' in \emph{CVPR}, 2022.

\bibitem{languagetask1}
A.~Fukui, D.~H. Park, D.~Yang, A.~Rohrbach, T.~Darrell, and M.~Rohrbach,
  ``Multimodal compact bilinear pooling for visual question answering and
  visual grounding,'' in \emph{EMNLP}, 2016.

\bibitem{languagetask2}
J.-H. Kim, J.~Jun, and B.-T. Zhang, ``Bilinear attention networks,'' in
  \emph{NIPS}, 2018.

\bibitem{languagetask3}
P.~Anderson, X.~He, C.~Buehler, D.~Teney, M.~Johnson, S.~Gould, and L.~Zhang,
  ``Bottom-up and top-down attention for image captioning and visual question
  answering,'' in \emph{CVPR}, 2018.

\bibitem{coca}
J.~Yu, Z.~Wang, V.~Vasudevan, L.~Yeung, M.~Seyedhosseini, and Y.~Wu, ``Coca:
  Contrastive captioners are image-text foundation models,'' \emph{arXiv},
  2022.

\bibitem{imgcaption1}
R.~Mokady, A.~Hertz, and A.~H. Bermano, ``Clipcap: Clip prefix for image
  captioning,'' \emph{arXiv}, 2021.

\bibitem{imgcaption2}
M.~Barraco, M.~Cornia, S.~Cascianelli, L.~Baraldi, and R.~Cucchiara, ``The
  unreasonable effectiveness of clip features for image captioning: an
  experimental analysis,'' in \emph{CVPR}, 2022.

\bibitem{visground1}
L.~H. Li, P.~Zhang, H.~Zhang, J.~Yang, C.~Li, Y.~Zhong, L.~Wang, L.~Yuan,
  L.~Zhang, J.-N. Hwang \emph{et~al.}, ``Grounded language-image
  pre-training,'' in \emph{CVPR}, 2022.

\bibitem{visground2}
G.~A. Sigurdsson, J.-B. Alayrac, A.~Nematzadeh, L.~Smaira, M.~Malinowski,
  J.~Carreira, P.~Blunsom, and A.~Zisserman, ``Visual grounding in video for
  unsupervised word translation,'' in \emph{CVPR}, 2020.

\bibitem{vlclassification}
A.~Radford, J.~W. Kim, C.~Hallacy, A.~Ramesh, G.~Goh, S.~Agarwal, G.~Sastry,
  A.~Askell, P.~Mishkin, J.~Clark \emph{et~al.}, ``Learning transferable visual
  models from natural language supervision,'' in \emph{ICML}, 2021.

\bibitem{vldetection}
A.~Kamath, M.~Singh, Y.~LeCun, G.~Synnaeve, I.~Misra, and N.~Carion,
  ``Mdetr-modulated detection for end-to-end multi-modal understanding,'' in
  \emph{ICCV}, 2021.

\bibitem{vlsegmentation}
Z.~Yang, J.~Wang, Y.~Tang, K.~Chen, H.~Zhao, and P.~H. Torr, ``Lavt:
  Language-aware vision transformer for referring image segmentation,'' in
  \emph{CVPR}, 2022.

\bibitem{DARTS}
H.~Liu, K.~Simonyan, and Y.~Yang, ``Darts: Differentiable architecture
  search,'' in \emph{ICLR}, 2019.

\bibitem{spos}
Z.~Guo, X.~Zhang, H.~Mu, W.~Heng, Z.~Liu, Y.~Wei, and J.~Sun, ``Single path
  one-shot neural architecture search with uniform sampling,'' in \emph{ECCV},
  2020.

\bibitem{yan2021lighttrack}
B.~Yan, H.~Peng, K.~Wu, D.~Wang, J.~Fu, and H.~Lu, ``Lighttrack: Finding
  lightweight neural networks for object tracking via one-shot architecture
  search,'' in \emph{CVPR}, 2021.

\bibitem{AutoMatch}
Z.~Zhang, Y.~Liu, X.~Wang, B.~Li, and W.~Hu, ``Learn to match: Automatic
  matching network design for visual tracking,'' \emph{ICCV}, 2021.

\bibitem{youtube}
E.~Real, J.~Shlens, S.~Mazzocchi, X.~Pan, and V.~Vanhoucke,
  ``Youtube-boundingboxes: A large high-precision human-annotated data set for
  object detection in video,'' in \emph{CVPR}, 2017.

\bibitem{bert}
J.~Devlin, M.-W. Chang, K.~Lee, and K.~Toutanova, ``Bert: Pre-training of deep
  bidirectional transformers for language understanding,'' \emph{arXiv}, 2018.

\bibitem{channelattention}
Z.~Shen, M.~Zhang, H.~Zhao, S.~Yi, and H.~Li, ``Efficient attention: Attention
  with linear complexities,'' in \emph{WACV}, 2021.

\bibitem{maskrcnn}
K.~He, G.~Gkioxari, P.~Doll{\'a}r, and R.~Girshick, ``Mask r-cnn,'' in
  \emph{ICCV}, 2017.

\bibitem{shufflenet}
X.~Zhang, X.~Zhou, M.~Lin, and J.~Sun, ``Shufflenet: An extremely efficient
  convolutional neural network for mobile devices,'' in \emph{CVPR}, 2018.

\bibitem{Prove}
Y.~Huang, C.~Du, Z.~Xue, X.~Chen, H.~Zhao, and L.~Huang, ``What makes
  multi-modal learning better than single (provably),'' \emph{NeurIPS}, 2021.

\bibitem{ERM}
A.~Singh, ``Foundations of machine learning,'' \emph{Available at SSRN
  3399990}, 2019.

\bibitem{risk1}
M.~R. Amini, N.~Usunier, and C.~Goutte, ``Learning from multiple partially
  observed views-an application to multilingual text categorization,''
  \emph{NIPS}, 2009.

\bibitem{risk2}
N.~Tripuraneni, M.~Jordan, and C.~Jin, ``On the theory of transfer learning:
  The importance of task diversity,'' \emph{NeurIPS}, 2020.

\bibitem{Rademacher}
P.~L. Bartlett and S.~Mendelson, ``Rademacher and gaussian complexities: Risk
  bounds and structural results,'' \emph{Journal of Machine Learning Research},
  2002.

\bibitem{kys}
G.~Bhat, M.~Danelljan, L.~V. Gool, and R.~Timofte, ``Know your surroundings:
  Exploiting scene information for object tracking,'' in \emph{ECCV}, 2020.

\bibitem{siamrcnn}
P.~Voigtlaender, J.~Luiten, P.~H. Torr, and B.~Leibe, ``Siam r-cnn: Visual
  tracking by re-detection,'' in \emph{CVPR}, 2020.

\bibitem{swintrack}
L.~Lin, H.~Fan, Z.~Zhang, Y.~Xu, and H.~Ling, ``Swintrack: A simple and strong
  baseline for transformer tracking,'' \emph{NeurIPS}, 2022.

\bibitem{COCO}
T.-Y. Lin, M.~Maire, S.~J. Belongie, L.~D. Bourdev, R.~B. Girshick, J.~Hays,
  P.~Perona, D.~Ramanan, P.~Doll{\'a}r, and C.~L. Zitnick, ``{Microsoft COCO}:
  Common objects in context,'' in \emph{ECCV}, 2014.

\bibitem{resnet}
K.~He, X.~Zhang, S.~Ren, and J.~Sun, ``Deep residual learning for image
  recognition,'' in \emph{CVPR}, 2016.

\bibitem{rebuttal_1}
H.-S. Chang, E.~Learned-Miller, and A.~McCallum, ``Active bias: Training more
  accurate neural networks by emphasizing high variance samples,'' \emph{NIPS},
  2017.

\bibitem{rebuttal_2}
B.~Kim, H.~Kim, K.~Kim, S.~Kim, and J.~Kim, ``Learning not to learn: Training
  deep neural networks with biased data,'' in \emph{CVPR}, 2019.

\bibitem{rebuttal_3}
N.~Mehrabi, F.~Morstatter, N.~Saxena, K.~Lerman, and A.~Galstyan, ``A survey on
  bias and fairness in machine learning,'' \emph{ACM Computing Surveys}, 2021.

\bibitem{gpt}
A.~Radford, J.~Wu, R.~Child, D.~Luan, D.~Amodei, I.~Sutskever \emph{et~al.},
  ``Language models are unsupervised multitask learners,'' \emph{OpenAI Blog},
  2019.

\bibitem{gradcam}
R.~R. Selvaraju, M.~Cogswell, A.~Das, R.~Vedantam, D.~Parikh, and D.~Batra,
  ``Grad-cam: Visual explanations from deep networks via gradient-based
  localization,'' in \emph{ICCV}, 2017.

\end{thebibliography}

%





\end{document}